\documentclass[11pt]{article}

\usepackage{acl}

\usepackage{times}
\usepackage{latexsym}

\usepackage[T1]{fontenc}

\usepackage[utf8]{inputenc}

\usepackage{microtype}

\usepackage{inconsolata}
\usepackage{multirow}
\usepackage{graphicx, subfigure}
\graphicspath{{img/}}
\usepackage{xcolor}

\usepackage{tipa}
\usepackage{amsmath}
\usepackage{amssymb}
\usepackage{booktabs}
\usepackage{alphalph}

\title{Hard to Be Heard: Phoneme-Level ASR Analysis of \\ Phonologically Complex, Low-Resource Endangered Languages}

\author{
  \textbf{V.S.D.S. Mahesh Akavarapu\textsuperscript{1}},
  \textbf{Michael Daniel\textsuperscript{2}},
  \textbf{Gerhard J\"{a}ger\textsuperscript{1}}
\\
  \textsuperscript{1}University of T\"{u}bingen,
  \textsuperscript{2}University of Jena
\\
\texttt{mahesh.akavarapu@uni-tuebingen.de, misha.daniel@gmail.com,}\\
\texttt{gerhard.jaeger@uni-tuebingen.de} \\
}

\begin{document}
\maketitle
\begin{abstract}

We present a phoneme-level analysis of automatic speech recognition (ASR) for two low-resourced and phonologically complex East Caucasian languages, Archi and Rutul, based on curated and standardized speech–transcript resources totaling approximately 50 minutes and 1 hour 20 minutes of audio, respectively. Existing recordings and transcriptions are consolidated and processed into a form suitable for ASR training and evaluation. We evaluate several state-of-the-art audio and audio--language models, including wav2vec2, Whisper, and Qwen2-Audio. For wav2vec2, we introduce a language-specific phoneme vocabulary with heuristic output-layer initialization, which yields consistent improvements and achieves performance comparable to or exceeding Whisper in these extremely low-resource settings. Beyond standard word and character error rates, we conduct a detailed phoneme-level error analysis. We find that phoneme recognition accuracy strongly correlates with training frequency, exhibiting a characteristic sigmoid-shaped learning curve. For Archi, this relationship partially breaks for Whisper, pointing to model-specific generalization effects beyond what is predicted by training frequency. Overall, our results indicate that many errors attributed to phonological complexity are better explained by data scarcity. These findings demonstrate the value of phoneme-level evaluation for understanding ASR behavior in low-resource, typologically complex languages.

\end{abstract}

\section{Introduction}
\begin{figure}
    \centering
    \includegraphics[width=0.8\columnwidth]{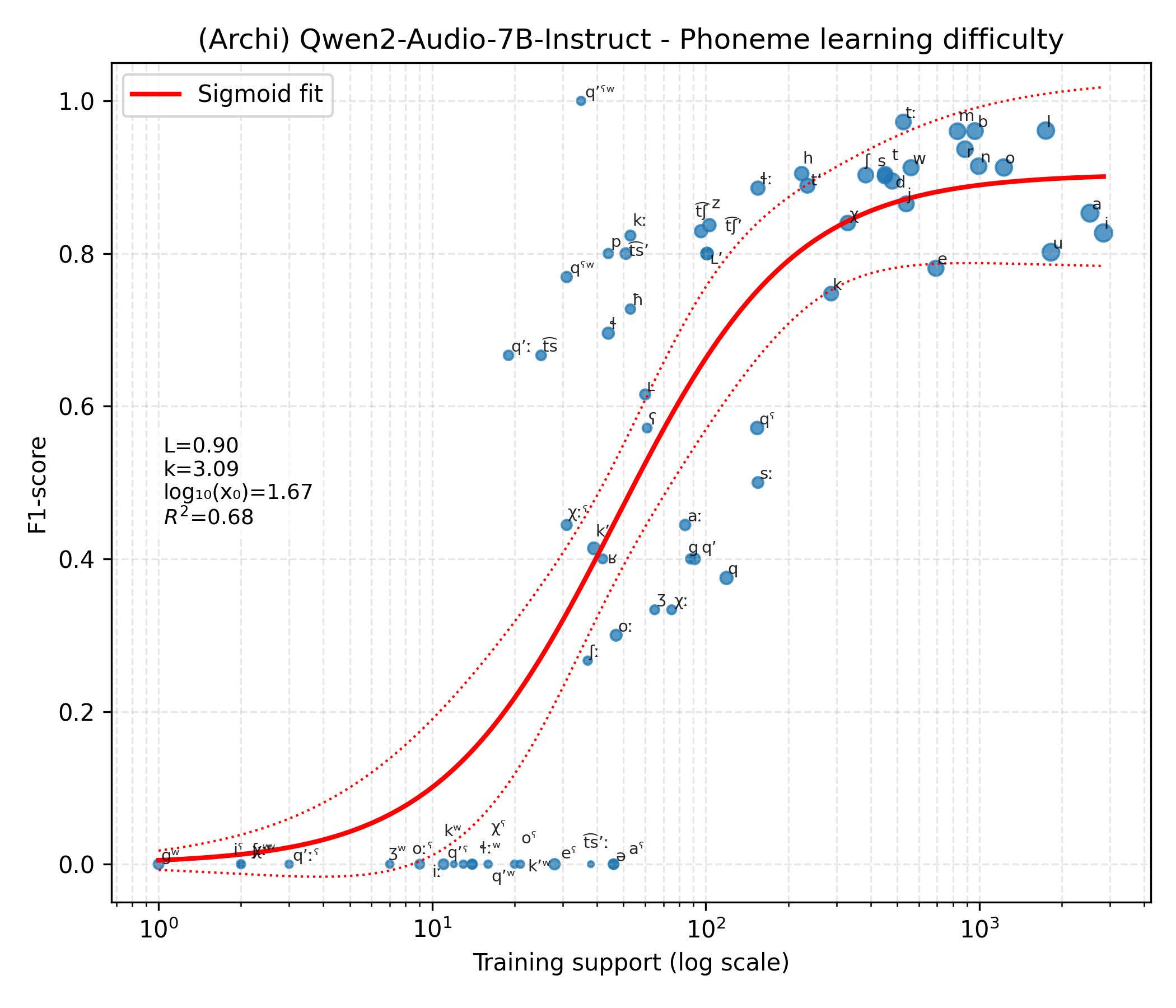}
    \caption{Phoneme-level F1-score as a function of (log) training frequency for one model–language pair, illustrating a characteristic sigmoid-shaped learning trend.}
    \label{fig:intro-sigmoid}
\end{figure}
Rutul and Archi are two East Caucasian languages with exceptionally rich sound systems that pose major challenges for modern automatic speech recognition (ASR). Archi, in particular, exhibits an unusually complex phonological system with 16 vowel phonemes and—depending on analytical assumptions—between 73 and 81 consonant phonemes, making it one of the largest non-click consonant inventories described to date. Rutul, represented here by its Kina variety, likewise features a large consonant inventory and special articulations such as pharyngealization. Both languages are highly endangered: Archi has only a few thousand speakers and is considered among the most severely endangered languages of Russia \citep{kibrik1977opyt, chumakina2007dictionary}, while Rutul is classified as definitely endangered with an estimated $\sim$30{,}000 speakers \citep{alekseevaetal2024}. Precise counts are difficult due to the geographic dispersion of speaker communities across Daghestan. Crucially, there are no established ASR benchmarks or standardized resources for either language.

Prior ASR research has largely focused on word- or character-level evaluation and has rarely examined phoneme-level behavior in typologically extreme languages, with only limited exceptions \citep[e.g.,][]{li-niehues-2025-enhance}. As a result, it remains unclear whether errors attributed to phonological complexity reflect intrinsic difficulty or simply data scarcity.

To address these gaps, we curate and standardize speech–transcript resources for Archi and Kina Rutul, consolidating material from linguistic documentation projects \citep{kibrik2007Archi, alekseevaetal2024} into a form suitable for ASR training and evaluation. The two corpora differ markedly in recording conditions: Kina Rutul consists primarily of spontaneous speech recorded in relatively noisy environments, whereas the Archi data comprise read speech produced by trained speakers under controlled conditions. Despite these differences, we observe broadly comparable trends across the two languages.

We fine-tune and evaluate several state-of-the-art ASR models on these resources, building on recent work showing that useful phonetic recognizers can be obtained from minutes of speech when combined with multilingual pretraining \citep{boulianne-2022-phoneme}. By benchmarking multiple architectures under identical conditions, we assess how current ASR systems handle languages with extreme phonological inventories.

Our phoneme-level analysis reveals a consistent pattern across most models and both languages: phoneme recognition accuracy, measured by F1 score, increases sigmoidally with the logarithm of training frequency (Figure~\ref{fig:intro-sigmoid}). Performance is near zero for very rare phonemes, rises sharply once sufficient examples are observed, and saturates for frequent ones. This mirrors frequency effects reported in cognitive and statistical models of language processing, where logistic functions of log-frequency capture performance trends \citep{heitmeier2024frequency}.

\paragraph{Contributions}
We summarize our main contributions as follows:
\begin{enumerate}
    \item We curate speech–transcript resources for Archi and Kina Rutul, enabling systematic ASR benchmarking for two East Caucasian languages previously lacking such resources.
    \item We benchmark multiple state-of-the-art ASR architectures including our introduction, a heuristic initialization trick by averaging of CTC based architecture, under extremely low-resource conditions, highlighting the advantages of speech-specialized models.
    \item We provide a detailed phoneme-level error analysis, revealing a robust sigmoid-shaped relationship between phoneme recognition accuracy and training frequency.
\end{enumerate}

\section{Related work}

Approaches to low-resource ASR primarily rely on cross-lingual transfer through multilingual pretraining, often using shared phonemic representations such as IPA or standardized phoneme sets derived from high-resource languages \citep{siminyu2021phoneme, li2022asr2k, taguchi2023universal, pratap2024scaling}. Another common strategy is augmenting end-to-end ASR systems with external $n$-gram language models to improve decoding in data-scarce settings \citep{xu2022simple, guillaume2022plugging, li-niehues-2025-enhance}. More recently, heuristic initialization of model layers---particularly output embeddings for unseen phonemes---has been explored to better transfer phonetic knowledge across languages \citep{yusuyin2025whistle}. Parallel to these developments, several works combine pretrained speech encoders with Large Language Models (LLMs) \citep{brown2020language, bai2023qwen}, either keeping the language model frozen \citep{fong2025speech} or adapting it using parameter-efficient techniques such as Low-Rank Adaptation (LoRA) \citep{hu2022lora, geng2025osum}.

Despite extensive linguistic documentation, computational resources and ASR studies for East Caucasian languages remain limited. There is no prior ASR work for Archi or Rutul, and only isolated efforts exist for other languages in the family, typically with very small datasets and without phoneme-level evaluation \citep{li2025context}.

Finally, while ASR performance is sometimes analyzed by broad phonological features/categories (e.g., tone, nasality, vowel length) \citep{liang-levow-2025-breaking} or such features are utilized in building ASR systems \citep{arora2018phonological}, detailed phoneme-level analyses remain rare, especially for phonologically complex and low-resource languages.

\section{Datasets}
\label{sec:datasets}

We work with manually transcribed speech data for two endangered East Caucasian languages from the Lezgic branch: Archi and Rutul. Our contribution lies in curating, consolidating, and standardizing existing speech–transcript resources into a form suitable for ASR training and evaluation. The resulting benchmark consists of:
\\\textbf{Archi:} approximately 45 minutes of training data (545 sentences) and 7 minutes of test data (100 sentences), derived from materials documented in \citet{kibrik2007Archi}. The recordings contain speech which was read out under controlled conditions.
\\\textbf{Kina Rutul:} approximately 75 minutes of training data (1{,}388 sentences) and 7 minutes of test data (90 sentences), based on documentation work reported in \citet{alekseevaetal2024}. The recordings contains spotaneous speech often recorded in noisy environments.

\begin{table}[]
\resizebox{\columnwidth}{!}{
\begin{tabular}{l|l|c|c|cc} \toprule
\multicolumn{1}{c}{\multirow{2}{*}{\textbf{Lang.}}} & \multicolumn{1}{c}{\multirow{2}{*}{\textbf{Split}}} & \multicolumn{1}{c}{\multirow{2}{*}{\textbf{Size}}} & \multicolumn{3}{|c}{\textbf{Vocabulary Sizes}}                                                                           \\ \cmidrule{4-6}
\multicolumn{1}{c}{}                                & \multicolumn{1}{c}{}                                & \multicolumn{1}{c}{}                               & \multicolumn{1}{|c}{\textbf{Words}} & \multicolumn{1}{|c}{\textbf{Phonemes}} & \multicolumn{1}{c}{\textbf{Composites}} \\ \midrule
\multirow{2}{*}{Archi}                              & Train                                               & 545 / 45m                                               & 1445                               & 85                                  & 50   (59\%)                                   \\
                                                    & Test                                                & 100    / 7m                                            & 394                                & 70                                  & 37    (53\%)                                  \\ \midrule
\multirow{2}{*}{Rutul}                              & Train                                               & 1388  / 75m                                             & 4866                               & 78                                 & 40      (51\%)                                \\
                                                    & Test                                                & 90   / 7m                                              & 441                                & 58                                  & 21   (36\%)       \\ \bottomrule                           
\end{tabular}
}
\caption{Dataset statistics --- split-wise number of sentences, total length in minutes (m), number of unique words, phones and composite (complex) phones.}
\label{tab:datasets}
\end{table}

Detailed split-wise statistics are provided in Table \ref{tab:datasets}. Composite phonemes are those with a diacritic such as \textipa{\super Q}, \textsuperscript{w}, ' or \textipa{:}. All possible phonemes per language are listed in the Appendix~\ref{app:cyrlmap}.

The original annotations differ substantially across the two corpora. Kina Rutul recordings are aligned with sentence-level annotations in \texttt{Praat} TextGrid format, while Archi recordings are aligned using \texttt{ELAN}. In both cases, the speech segments are segmented at the sentence level. The transcriptions themselves are heterogeneous, combining IPA symbols, romanized conventions (e.g. \v{s} for \textipa{S}), and occasional Cyrillic-based notation (e.g.\ \verb+|+ to mark pharyngealization i.e., \textipa{\super Q}). As part of the curation process, we normalize these annotations into consistent, sentence-level IPA transcriptions paired with the corresponding audio. We adopt IPA to facilitate transfer from multilingual pretrained ASR models.

In terms of speaker coverage, the Kina Rutul data include recordings from approximately 15 adult speakers (with a slight predominance of female speakers), while the Archi data consist of read speech produced by two trained female speakers. Although the corpora are limited in size, their curation enables controlled phoneme-level ASR experiments that were previously not feasible.

\section{Models}

We evaluated the following model families and their modifications:

\paragraph{wav2vec2-large-ipa} \citep{taguchi2023universal} This is a wav2vec~2.0 model \citep{baevski2020wav2vec} fine-tuned for ASR on connectionist temporal classification (CTC) \citep{graves2006connectionist} using IPA transcriptions from multiple languages, and is therefore suitable for zero-shot evaluation on unseen languages (\textbf{-zs} suffixed).

\paragraph{w2v2l-custom}
For wav2vec2-large-ipa, we additionally define a language-specific phoneme vocabulary derived from the IPA transcriptions. Composite phonemes (e.g., labialized or pharyngealized consonants) are mapped to a reduced vocabulary reflecting the actual phonemic contrasts of each language. For example, the phoneme k\super w is tokenized by the wav2vec2-large-ipa tokenizer as the sequence `k',`\super w', whereas in this model it is represented as a single token. We experiment with two initialization strategies for the output layer of wav2vec2, denoted by suffixes; without any suffix, the final layer is randomly initialized.

\paragraph{w2v2l-custom-avg}
Columns corresponding to composite phonemes are initialized by averaging the pretrained parameters of their component IPA symbols. Let the weights and biases of the output layer be $W \in \mathbb{R}^{d \times |V|}$ and $b \in \mathbb{R}^{|V|}$, where $d$ is the hidden dimension and $V$ is the reduced, language-specific vocabulary. Let the pretrained weights, biases, and vocabulary be denoted by $W^{\text{old}}$, $b^{\text{old}}$, and $V^{\text{old}}$, respectively. For a phoneme indexed by $i$ in $V^{\text{new}}$ that is composed of symbols indexed by $i_1, \ldots, i_k$ in $V^{\text{old}}$, the parameters are initialized as:
\[
W_{*i} = \frac{1}{k} \sum_{j=1}^k W^{\text{old}}_{*i_j}
\quad ; \quad
b_i = \frac{1}{k} \sum_{j=1}^k b^{\text{old}}_{i_j}.
\]
Averaging is our novel step. We also perform zero-shot evaluation with this model enabled by non-random initialization, with suffix \textbf{-zs} in the name.

\paragraph{w2v2l-custom-cpy1}
In this variant, the output-layer parameters corresponding to the base phoneme (i.e., without diacritics) are copied directly, rather than averaged, following a strategy similar to \citet{yusuyin2025whistle}.

\paragraph{w2v2l-custom-avg-lm}
Inspired from \citet{xu2022simple, guillaume2022plugging, li-niehues-2025-enhance}, we incorporate a word-level $n$-gram language model (LM) with $n=3$ on top of w2v2l-custom-avg to reduce word error rate. This differs from previous works slightly as the latter incorporate character/phoneme $n$-gram, which we did not find improving performance. The CTC output sequence includes a word-separator token `\texttt{|}', which deterministically segments the phoneme sequence $X$ into words $w_1(X), \ldots, w_{m(X)}(X)$. The vocabulary of LM consists of words occurring in the training transcripts. The mapping from segmented phoneme sequences to LM tokens is handled directly in a dictionary-like manner. During decoding, the following objective is maximized over a phoneme sequence $X = x_1, \ldots, x_l$ and its corresponding word sequence $w_1(X), \ldots, w_{m(X)}(X)$:
\[
\max_{X = x_1, \ldots, x_l}
\sum_{i=1}^l \log p_{\text{ctc}}(x_i) + \beta \cdot m(X) +
\]
\vspace*{-0.4cm}
\[
\alpha\sum_{i=n+1}^{m(X)} \log p_{\text{lm}}\!\left(w_i(X) \mid w_{i-1}(X), \ldots, w_{i-n}(X)\right)
\]
where $p_{\text{ctc}}$ denotes the CTC softmax probabilities produced by wav2vec2, $p_{\text{lm}}$ is the probability from the $n$-gram language model, and $\alpha$ and $\beta$ are tunable hyperparameters. The optimal sequence is approximated using beam search. We use KenLM \citep{heafield-2011-kenlm} for $n$-gram language modeling. 

\paragraph{whisper-large-v3} \citep{radford2023robust} This is a multilingual pretrained encoder--decoder model that supports IPA output. Due to its inherent subword tokenizer (as well as those of the models described below), vocabulary reduction and output-layer reinitialization are not straightforward; therefore, the model is fine-tuned without modifying its pretrained decoder vocabulary.

\paragraph{Qwen2-Audio-7B-Instruct} \citep{chu2024qwen2} combines an audio encoder (initialized from whisper-large) with a large language model (LLM) (Qwen2; \citet{team2024qwen2}), aligning the encoder outputs with the input representation space of the LLM.

\paragraph{Qwen2.5-Omni-7B} \citep{xu2025qwen2} This model supports text, vision, and audio inputs and can generate spoken responses. In this work, we restrict the model to audio input and text output. For both Qwen-based models, the audio encoder is fine-tuned, while the LLM is fine-tuned using LoRA.

\paragraph{gpt-4o-transcribe} \citep{hurst2024gpt} This model is accessed via the OpenAI API and is therefore used without fine-tuning. The model predominantly outputs Cyrillic script. Thus, we explicitly prompt it to transcribe in Cyrillic. 

Since the conversion between Cyrillic, the official script for these languages (only introduced recently), and IPA is deterministic (see Appendix \ref{app:cyrlmap} for mappings), we additionally fine-tune generative models to output Cyrillic. These variants are denoted as \textbf{whisper-large-v3-cyrl}, \textbf{Qwen2-Audio-7B-Instruct-cyrl}, and \textbf{Qwen2.5-Omni-7B-cyrl}. Evaluations are carried out after converting to IPA.

\section{Experiments}
\subsection{Implementation Details}

We create a validation set by holding out 5\% of the training sentences. The vocabulary sizes of w2v2l-custom* models are 90 and 84 respectively for Archi and Rutul including special characters. Learning rates are set to $3\times10^{-5}$ for CTC-based models and $5\times10^{-6}$ for Whisper and LLM-coupled audio encoders. All models are optimized using Adam \citep{kingma2014adam} with a weight decay of 0.01 \citep{loshchilov2017decoupled}. For Qwen-based models, the LoRA parameters are rank $r=16$, scaling factor $\alpha=32$, and dropout 0.05 (similar to \citet{geng2025osum}). LoRA is applied to all linear layers and optimized using Adam with a learning rate of $1\times10^{-4}$ and no weight decay. The number of trainable and total parameters per model are listed in Table \ref{tab:modelres}. CTC-based models are fine-tuned for 30 epochs, Whisper for 10 epochs, and Qwen-based models for 6 epochs. All models use an effective batch size of 16 via gradient accumulation. CTC-based models are trained on two NVIDIA RTX 2080 GPUs (11\,GB $\times 2$), while larger models are trained on a single NVIDIA H100 (80\,GB).

For the LLM-based models, we use the following prompt, refined through a small number of manual trials with gpt-4o-transcribe:

\texttt{``Transcribe the audio in <lang> (a Northeast Caucasian language) into 
<IPA (International Phonetic Alphabet) | Cyrillic>. Do not translate, interpret, or add punctuation. Output only the phonetic transcription.''}

The 3-gram language model coupled with CTC uses $\alpha=\beta=0.3$, tuned on the validation sets with a beam size of 10 (restricted to this value for efficiency). The code\footnote{\url{https://github.com/mahesh-ak/north_caucasian_asr}} and the datasets\footnote{\url{https://huggingface.co/datasets/mahesh27/archi_rutul_asr}} are publicly available.

\subsection{Evaluation metrics}
We evaluate on standard metrics for ASR --- word, character and phoneme error rates, respectively WER, CER and PER which are normalized edit distances respectively at levels of words, characters and phonemes. We further store the number of edits --- insertions (I), deletions (D) and substitutions (S) --- along with true positives (N) for each phoneme to compute phoneme-level precision (pr), recall (re) and F1 scores:

\begin{equation*}
\resizebox{\columnwidth}{!}{
$\text{pr} = \frac{N}{N + S + I}\text{ ; }\text{re} = \frac{N}{N + S + D} ;\text{F1} = \frac{2\cdot\text{pr}\cdot\text{re}}{\text{pr}+\text{re}}$
}
\end{equation*}

\subsection{Modeling of Phoneme Frequency Effects}
\label{subsec:exp-sigmoid}
To analyze the relationship between phoneme recognition performance and data availability, we model phoneme-level F1 scores as a function of log training frequency using a logistic function:
\[f(x) = \frac{L}{1 + \exp(-k(x - x_0))} \]
where $x = \log_{10}(\text{training frequency})$, $L$ denotes the asymptotic F1 score, $k$ controls the slope, and $x_0$ is the midpoint of the transition. We emphasize that the logistic form is used as a descriptive parametric summary of the observed nonlinear trend, rather than as a theoretical assumption. Alternative shapes (e.g., piecewise-linear or threshold-like behavior) may also fit the data; our goal is to capture a consistent frequency–accuracy scaling pattern and estimate interpretable midpoints, rather than to commit to a specific functional form.

Parameters are estimated via non-linear least squares using the Levenberg-Marquardt algorithm \citep{marquardt1963algorithm}. Model fit is quantified using the coefficient of determination ($R^2$) between observed and predicted F1 scores. Uncertainty in the fitted curve is assessed using approximate 95\% confidence intervals derived via the Delta method \citep{van2000asymptotic}.

\section{Results}

\begin{table*}[t!]
\centering
\resizebox{0.7\textwidth}{!}{
\begin{tabular}{l|c|c|ccc|ccc} \toprule
    \multicolumn{3}{c}{}           & \multicolumn{3}{|c|}{\textbf{Archi}}               & \multicolumn{3}{c}{\textbf{Rutul}}               \\ \midrule
\textbf{Model}  & \textbf{Params.} & \textbf{Tunable}        & \textbf{WER}   & \textbf{CER}   & \textbf{PER}   & \textbf{WER}   & \textbf{CER}   & \textbf{PER}   \\ \midrule
wav2vec2-large-ipa-zs & 0.3B & -   & 1.000          & 0.593          & 0.606          & 1.000          & 0.656          & 0.660          \\
wav2vec2-large-ipa  & 0.3B &  0.3B    & 0.559          & 0.128          & 0.135          & 0.795          & 0.223          & 0.220          \\
w2v2l-custom-avg-zs & 0.3B &  -    & 1.000          & 0.544          & 0.558          & 1.000          & 0.563          & 0.571          \\
w2v2l-custom     & 0.3B &   0.3B      & 0.593          & 0.138          & 0.147          & 0.780          & 0.224          & 0.222          \\
w2v2l-custom-cpy1  & 0.3B &   0.3B    & 0.462          & 0.116          & 0.123          & 0.738          & 0.205          & 0.203          \\
w2v2l-custom-avg (ours)  & 0.3B &  0.3B     & 0.479          & 0.116          & 0.122          & 0.725          & \textbf{0.198} & \textbf{0.195} \\
w2v2l-custom-avg-lm  (ours) & 0.3B &  0.3B  & 0.465          & 0.116          & 0.122          & \textbf{0.697} & 0.206          & 0.206          \\
whisper-large-v3  & 1.5B &  1.5B      & \textbf{0.402} & \textbf{0.099} & \textbf{0.107} & 0.778          & 0.253          & 0.251          \\
whisper-large-v3-cyrl   & 1.5B & 1.5B      & 0.422          & 0.111          & 0.119          & 0.792          & 0.232          & 0.235          \\
Qwen2-Audio-7B-Instruct  & 8.4B &  0.7B    & 0.579          & 0.163          & 0.180           & 0.778          & 0.242          & 0.239          \\
Qwen2-Audio-7B-Instruct-cyrl & 8.4B &  0.7B  & 0.539          & 0.156          & 0.166          & 0.828          & 0.272          & 0.274          \\
Qwen2.5-Omni-7B      & 10.8B &    0.7B      & 0.705          & 0.184          & 0.199          & 0.852          & 0.263          & 0.257          \\
Qwen2.5-Omni-7B-cyrl   & 10.8B &  0.7B      & 0.904          & 0.295          & 0.291          & 0.904          & 0.295          & 0.291          \\
gpt-4o-transcribe    & - & -        & 0.982          & 0.435          & 0.436          & 0.994          & 0.519          & 0.514 \\ \bottomrule    
\end{tabular}
}
\caption{ASR Performance of models on Archi and Kina Rutul in terms of Word- (WER), Character- (CER) and Phoneme- (PER) error rates. Lower the error rates better the model. Best performances are in \textbf{bold}.}
\label{tab:modelres}
\end{table*}

\begin{table*}[t]
\centering
\resizebox{\textwidth}{!}{
\begin{tabular}{l|lllllllllllllllll|l} \toprule
\textbf{Complexity$\rightarrow$}                  & \textbf{1}     & \textbf{2}     & \textbf{2}     & \textbf{3}     & \textbf{3}     & \textbf{4}     & \textbf{3}     & \textbf{4}    & \textbf{2}     & \textbf{3}     & \textbf{3}     & \textbf{2}    & \textbf{3}   & \textbf{1}     & \textbf{2}     & \textbf{3}   & \textbf{2} &   \\ \midrule
\textbf{Model$\downarrow$}                        & \textbf{C}                     & \textbf{C\super w}                    & \textbf{C'}                    & \textbf{C'\super w}                   & \textbf{C'\textipa{:}}                   & \textbf{C'\textipa{: \super Q}}                  & \textbf{C'\textipa{\super Q}}                   & \textbf{C'\textipa{\super {Qw}}}                  & \textbf{C\textipa{:}}                    & \textbf{C\textipa{:}\super w}                   & \textbf{C\textipa{: \super Q}}                   & \textbf{C\textipa{\super Q}}                    & \textbf{C\textipa{\super {Qw}}}                   & \textbf{V}                     & \textbf{V\textipa{:}}                    & \textbf{V\textipa{:\super Q}}                   & \textbf{V\textipa{\super Q}}       & \textbf{$r$}             \\ \midrule
gpt-4o-transcribe            & 0.449                 & 0.4                   & 0.035                 & 0.0                   & 0.0                   & 0.0                   & 0.0                   & 0.0                   & 0.355                 & 0.0                   & 0.0                   & 0.0                   & 0.0                   & 0.525                 & 0.234                 & 0.0                   & 0.0                   & -0.76                                  \\
whisper-large-v3             & \textbf{0.894}        & 0.667                 & \textbf{0.823}        & 0.333                 & 0.5                   & 0.0                   & 0.0                   & 0.667                 & 0.803                 & 0.722                 & 0.769                 & 0.673                 & 0.545                 & 0.736                 & 0.577                 & \textbf{1.0}          & 0.317                 & -0.44                                  \\
whisper-large-v3-cyrl        & 0.802                 & \textbf{0.863}        & 0.78                  & \textbf{0.667}        & 1.0                   & 0.0                   & 0.0                   & \textbf{1.0}          & 0.817                 & 0.889                 & \textbf{0.833}        & 0.419                 & 0.667                 & 0.742                 & \textbf{0.583}        & 0.889                 & \textbf{0.442}        & -0.15                                  \\
Qwen2-Audio-7B-Instruct      & 0.764                 & 0.0                   & 0.684                 & 0.0                   & 0.333                 & 0.0                   & 0.0                   & \textbf{1.0}          & 0.63                  & 0.0                   & 0.444                 & 0.286                 & 0.769                 & 0.696                 & 0.248                 & 0.0                   & 0.0                   & -0.21                                  \\
Qwen2-Audio-7B-Instruct-cyrl & 0.735                 & 0.636                 & 0.615                 & 0.333                 & \textbf{0.625}        & 0.0                   & 0.222                 & \textbf{1.0}          & 0.583                 & 0.333                 & 0.4                   & 0.35                  & 0.8                   & 0.709                 & 0.484                 & 0.667                 & 0.156                 & -0.18                                  \\
wav2vec2-large-ipa           & 0.878                 & 0.667                 & 0.846                 & 0.333                 & 0.4                   & 0.0                   & \textbf{0.333}        & 0.8                   & \textbf{0.862}        & \textbf{0.933}        & 0.417                 & \textbf{0.87}         & 0.933                 & 0.765                 & 0.483                 & 0.857                 & 0.345                 & -0.38                                  \\
wav2vec2-large-ipa-zs        & 0.308                 & 0.0                   & 0.0                   & 0.0                   & 0.0                   & 0.0                   & 0.0                   & 0.0                   & 0.188                 & 0.0                   & 0.0                   & 0.0                   & 0.0                   & 0.47                  & 0.153                 & 0.0                   & 0.0                   & -0.71                                  \\
Qwen2.5-Omni-7B              & 0.728                 & 0.0                   & 0.53                  & 0.0                   & 0.222                 & \textbf{0.667}        & 0.0                   & 0.0                   & 0.632                 & 0.0                   & 0.0                   & 0.348                 & 0.4                   & 0.681                 & 0.366                 & 0.0                   & 0.0                   & -0.46                                  \\
Qwen2.5-Omni-7B-cyrl         & 0.697                 & 0.303                 & 0.368                 & 0.0                   & 0.0                   & 0.0                   & 0.0                   & 0.5                   & 0.489                 & 0.167                 & 0.0                   & 0.283                 & 0.364                 & 0.661                 & 0.412                 & 0.0                   & 0.0                   & -0.6                                   \\
w2v2l-custom-avg             & 0.867                 & 0.222                 & 0.778                 & 0.333                 & 0.417                 & 0.0                   & 0.0                   & 0.857                 & 0.753                 & 0.0                   & 0.462                 & 0.426                 & 0.933                 & \textbf{0.773}        & 0.389                 & 0.0                   & 0.192                 & -0.36                                  \\
w2v2l-custom-avg-zs          & 0.34                  & 0.0                   & 0.0                   & 0.0                   & 0.0                   & 0.0                   & 0.0                   & 0.0                   & 0.0                   & 0.0                   & 0.0                   & 0.0                   & 0.0                   & 0.489                 & 0.0                   & 0.0                   & 0.0                   & -0.65                                  \\
w2v2l-custom                 & 0.835                 & 0.0                   & 0.79                  & 0.25                  & 0.143                 & 0.0                   & 0.0                   & \textbf{1.0}          & 0.749                 & 0.0                   & 0.455                 & 0.417                 & 0.933                 & 0.71                  & 0.341                 & 0.0                   & 0.0                   & -0.24                                  \\
w2v2l-custom-avg-lm          & 0.88                  & 0.333                 & 0.809                 & 0.333                 & 0.364                 & 0.0                   & 0.0                   & 0.857                 & 0.752                 & 0.0                   & 0.462                 & 0.426                 & 0.933                 & \textbf{0.773}        & 0.413                 & 0.0                   & 0.192                 & -0.39                                  \\
w2v2l-custom-cpy1            & 0.878                 & 0.333                 & 0.791                 & 0.333                 & 0.417                 & \textbf{0.667}        & 0.0                   & 0.857                 & 0.841                 & 0.222                 & 0.5                   & 0.447                 & \textbf{1.0}          & 0.771                 & 0.481                 & \textbf{1.0}          & 0.354                 & -0.09     \\ \bottomrule
\end{tabular}
}
\caption{Phoneme category-wise averaged F1 scores and their Pearson's correlation coefficients $r$ with complexities (length) - Archi}
\label{tab:phonwise-archi}
\end{table*}

\begin{table*}[t]
\centering
\resizebox{0.7\textwidth}{!}{
\begin{tabular}{l|llllllllll|l} \toprule
\textbf{Complexity$\rightarrow$}          & \textbf{1}     & \textbf{2}     & \textbf{3}   & \textbf{2}     & \textbf{2}     & \textbf{3}   & \textbf{3}     & \textbf{1}     & \textbf{2}     & \textbf{2}  &  \\ \midrule
\textbf{Model$\downarrow$}               & \textbf{C}     & \textbf{C\super j}    & \textbf{C\super j \textipa{:}} & \textbf{C \super w}    & \textbf{C'}    & \textbf{C'\super j} & \textbf{C'\super w}   & \textbf{V}     & \textbf{V\textipa{:}}    & \textbf{V\textipa{\super Q}}   & \textbf{$r$}\\ \midrule
gpt-4o-transcribe            & 0.396                 & 0.0                   & 0.0                   & 0.1                   & 0.0                   & 0.0                   & 0.0                   & 0.457                 & 0.034                 & 0.0                   & -0.81                                  \\
whisper-large-v3             & 0.76                  & 0.429                 & 0.0                   & 0.31                  & 0.37                  & 0.0                   & 0.0                   & 0.715                 & 0.056                 & 0.165                 & -0.92                                  \\
whisper-large-v3-cyrl        & 0.721                 & 0.333                 & 0.0                   & 0.417                 & 0.0                   & 0.0                   & 0.0                   & 0.678                 & 0.133                 & 0.0                   & -0.84                                  \\
Qwen2-Audio-7B-Instruct      & 0.712                 & \textbf{0.611}        & 0.0                   & 0.42                  & 0.374                 & 0.0                   & 0.0                   & 0.68                  & 0.094                 & 0.151                 & -0.88                                  \\
Qwen2-Audio-7B-Instruct-cyrl & 0.683                 & 0.0                   & 0.0                   & 0.228                 & 0.0                   & 0.0                   & 0.0                   & 0.665                 & 0.097                 & 0.0                   & -0.82                                  \\
wav2vec2-large-ipa           & 0.797                 & 0.389                 & 0.0                   & \textbf{0.468}        & \textbf{0.586}        & 0.0                   & 0.0                   & 0.725                 & 0.038                 & 0.261                 & -0.89                                  \\
wav2vec2-large-ipa-zs        & 0.326                 & 0.0                   & 0.0                   & 0.0                   & 0.0                   & 0.0                   & 0.0                   & 0.309                 & 0.22                  & 0.0                   & -0.77                                  \\
Qwen2.5-Omni-7B              & 0.693                 & 0.278                 & 0.0                   & 0.28                  & 0.246                 & 0.0                   & 0.0                   & 0.642                 & 0.071                 & 0.024                 & -0.9                                   \\
Qwen2.5-Omni-7B-cyrl         & 0.686                 & 0.0                   & 0.0                   & 0.199                 & 0.0                   & 0.0                   & 0.0                   & 0.607                 & 0.107                 & 0.0                   & -0.82                                  \\
w2v2l-custom-avg             & 0.79                  & 0.222                 & 0.0                   & 0.36                  & 0.567                 & 0.0                   & 0.0                   & \textbf{0.748}        & 0.107                 & 0.315                 & -0.92                                  \\
w2v2l-custom-avg-zs          & 0.415                 & 0.0                   & 0.0                   & 0.0                   & 0.0                   & 0.0                   & 0.0                   & 0.331                 & 0.0                   & 0.0                   & -0.78                                  \\
w2v2l-custom                 & 0.757                 & 0.222                 & 0.0                   & 0.356                 & 0.479                 & 0.0                   & 0.0                   & 0.699                 & 0.133                 & 0.263                 & -0.94                                  \\
w2v2l-custom-avg-lm         & 0.789                 & 0.333                 & 0.0                   & 0.363                 & 0.567                 & 0.0                   & 0.0                   & 0.747                 & 0.074                 & 0.32                  & -0.92                                  \\
w2v2l-custom-cpy1            & \textbf{0.809}        & 0.317                 & 0.0                   & 0.389                 & 0.506                 & 0.0                   & \textbf{0.667}        & 0.732                 & \textbf{0.138}        & \textbf{0.35}         & -0.66                              \\ \bottomrule   
\end{tabular}
}
\caption{Phoneme category-wise averaged F1 scores and their Pearson's correlation coefficients $r$ with complexities (length) - Kina Rutul}
\label{tab:phonwise-rutul}
\end{table*}
\subsection{Overall Performance}

Overall results are summarized in Table~\ref{tab:modelres}, with statistical significance assessed using paired wilcoxon signed-rank tests (Appendix~\ref{app:pvals}). As expected, the zero-shot models perform poorly, yielding near-random WER for both languages, underscoring the extreme low-resource and phonologically complex nature of the tasks.

On \textbf{Archi}, whisper-large-v3 achieves the best overall performance. Nevertheless, the w2v2l-custom variants substantially improve over the base wav2vec2-large-ipa model: w2v2l-custom-avg (ours) reduce WER by more than 8 absolute points, with statistically significant gains in both WER and CER ($p < 0.05$) (w2v2l-custom-cpy1 doesn't always give significant gain). Adding a word-level $3$-gram language model (w2v2l-custom-avg-lm) yields only marginal additional improvements, consistent with the limited amount of training data. The best (lowest) WER is comparable to results reported for similarly low-resourced languages (e.g., \citet{li-niehues-2025-enhance}).

On \textbf{Rutul}, similar trends hold. The w2v2l-custom-avg model achieves the lowest CER and PER, while w2v2l-custom-avg-lm yields the best WER (0.697), with statistically significant improvements over wav2vec2-large-ipa. In contrast to Archi, whisper-large-v3 performs noticeably worse on Rutul, suggesting weaker transfer. This asymmetry is consistent with later phoneme-level analyses, where Whisper exhibits fewer deviations from frequency-driven learning on Rutul. We further analyze data quality–quantity trade-offs for Rutul in Appendix~\ref{app:rutulqvsq}.

Across both languages, large audio–language models (Qwen2-Audio and Qwen2.5-Omni) underperform relative to CTC-based models, even after fine-tuning. Their Cyrillic-output variants (-cyrl), corresponding to the recently standardized official script (see \S\ref{sec:datasets}), generally lag behind their IPA-output counterparts. 

Overall performance improves with increasing specialization toward speech, from multimodal Qwen2.5-Omni to audio-centric Qwen2-Audio and dedicated ASR models such as wav2vec2 and Whisper, where simple, linguistically informed adaptations of CTC-based models---language-specific phoneme vocabularies with heuristic initialization---can match or outperform substantially larger pretrained systems. The subsequent phoneme-level analysis clarifies how training frequency and pretraining jointly shape recognition performance.

\subsection{Phoneme-level Analysis}

Tables~\ref{tab:phonwise-archi} and~\ref{tab:phonwise-rutul} report category-wise F1 scores along with their correlation (Pearson's $r$) with phoneme complexity. We define phoneme complexity as the number of additional articulatory features (e.g., \textipa{\super Q}, \textsuperscript{w}, ', \textipa{:}) attached to a base segment, as indicated by IPA diacritics, plus 1 for the base segment. In IPA, using a diacritic usually indicates an additional articulation - such as {\super w} for labialization or a feature that distinguishes the phoneme from its more ``typologically common'' counterpart, as ' for non-pulmonic (ejective) consonant. Such additional features are a reasonable proxy for articulatory complexity as well as for typological markedness. Across both languages, simpler segments are consistently recognized better than marked categories including secondary articulations, length contrasts, and pharyngealized or labialized segments (also see the phonemes with least F1 in Appendix \ref{app:toperr}).

For Rutul, we observe a clear negative correlation between complexity and performance across models, indicating systematic degradation as grapheme length or articulatory complexity increases. In contrast, Archi shows only weak to moderate negative correlations, suggesting that performance degradation is less monotonic and more category-specific rather than strictly driven by length. This weaker trend motivates an alternative treatment and accordingly, in the following section we model recognition performance as a function of log training frequency using a sigmoid formulation, which is independent of phoneme complexity.

In both overall performance (Table~\ref{tab:modelres}) and at the phoneme level, zero-shot models---gpt-4o-transcribe, wav2vec2-large-ipa-zs, and w2v2l-custom-avg-zs---serve as baselines, illustrating that without language-specific pre-training it is nearly impossible to reliably recognize articulatorily complex phonemes.

\begin{figure*}[t]
    \centering
    \subfigure[]{\includegraphics[width=0.16\textwidth]{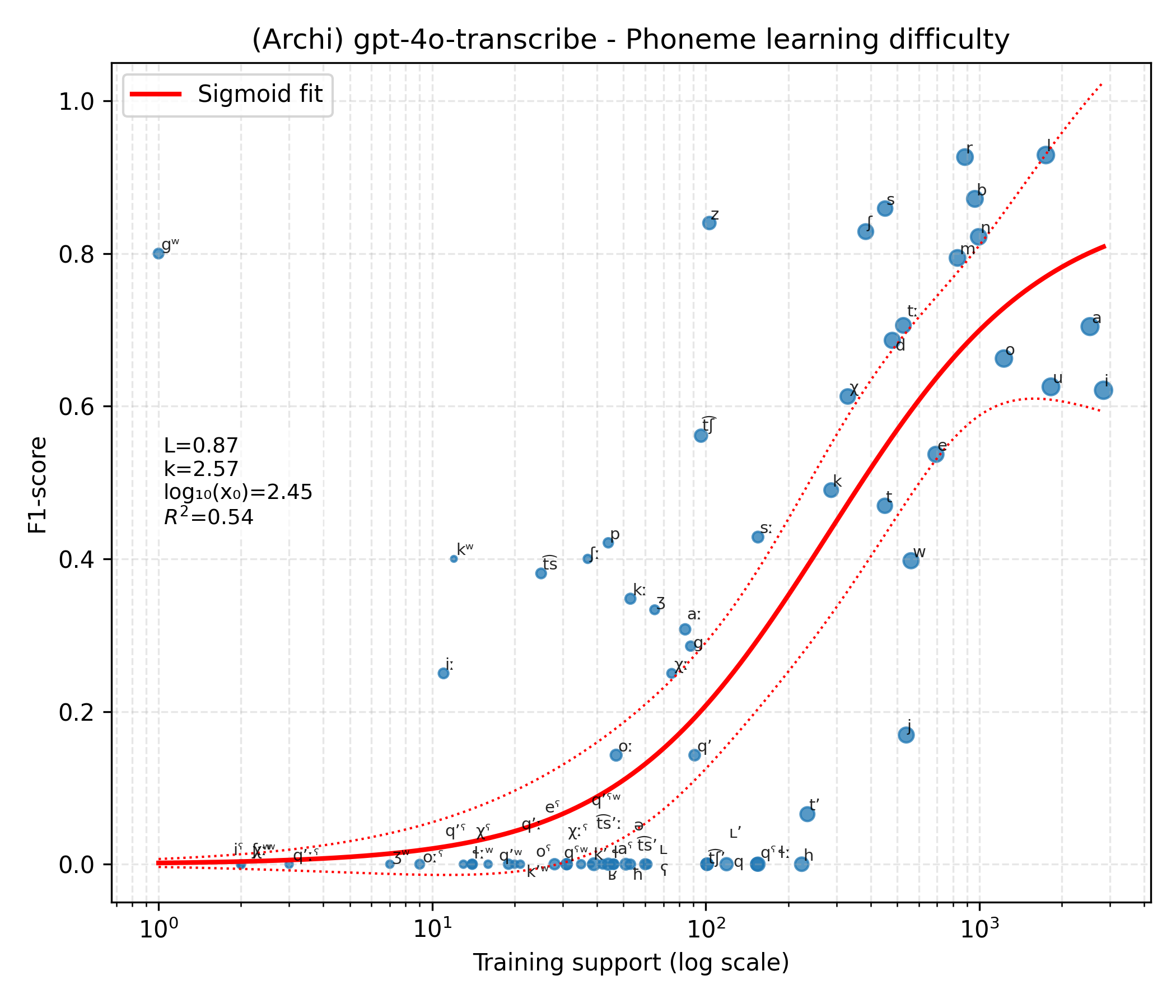}}
    \subfigure[]{\includegraphics[width=0.16\textwidth]{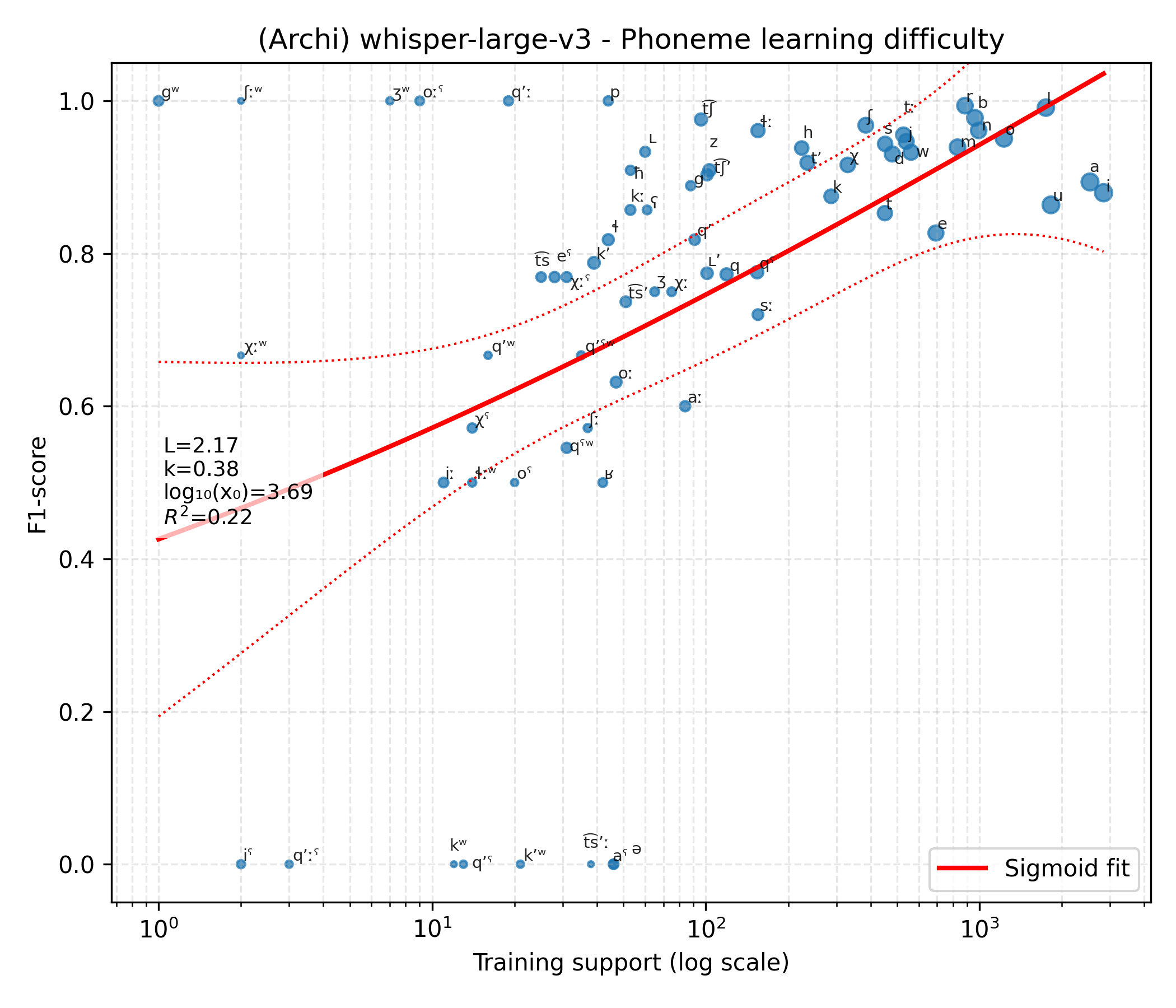}}
    \subfigure[]{\includegraphics[width=0.16\textwidth]{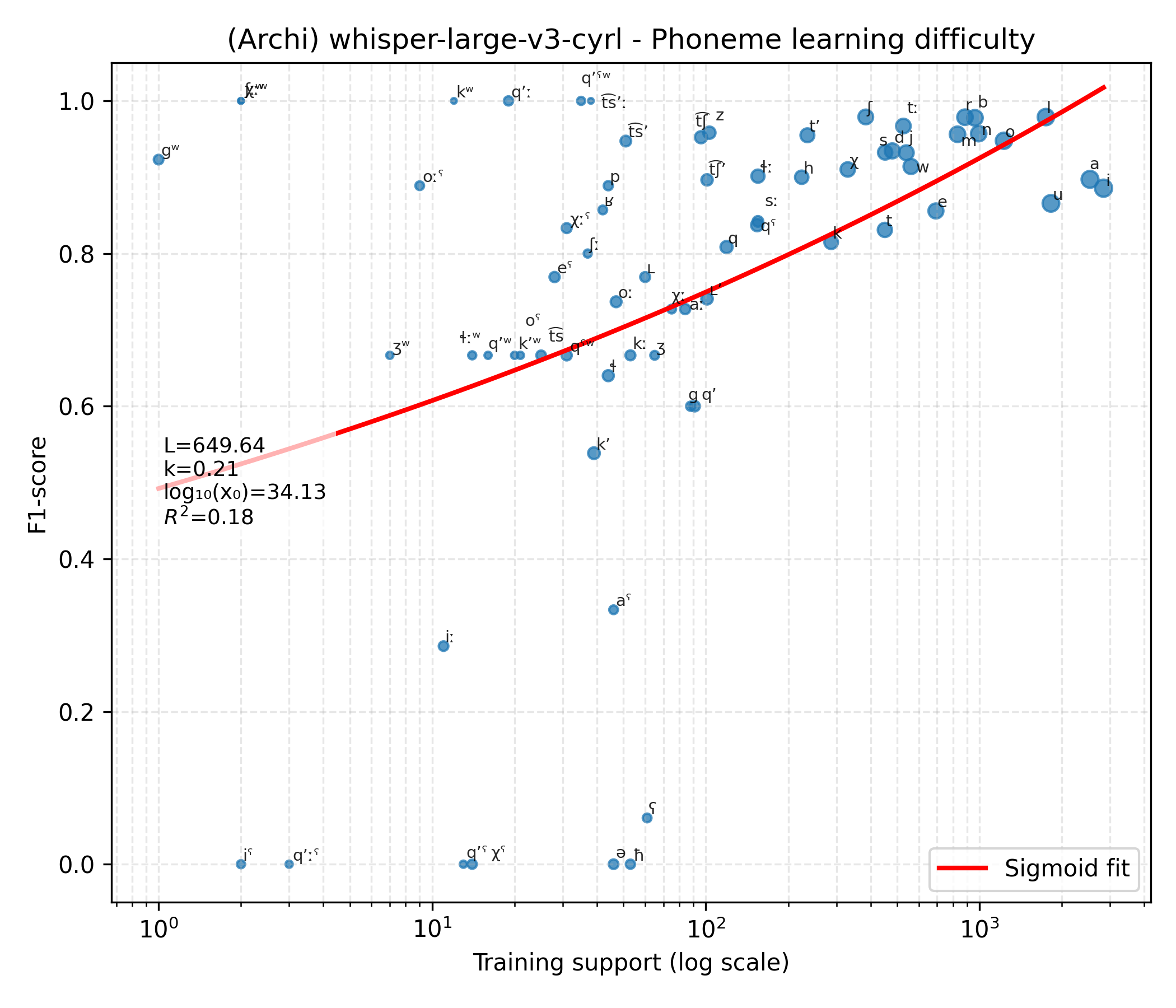}}
    \subfigure[]{\includegraphics[width=0.16\textwidth]{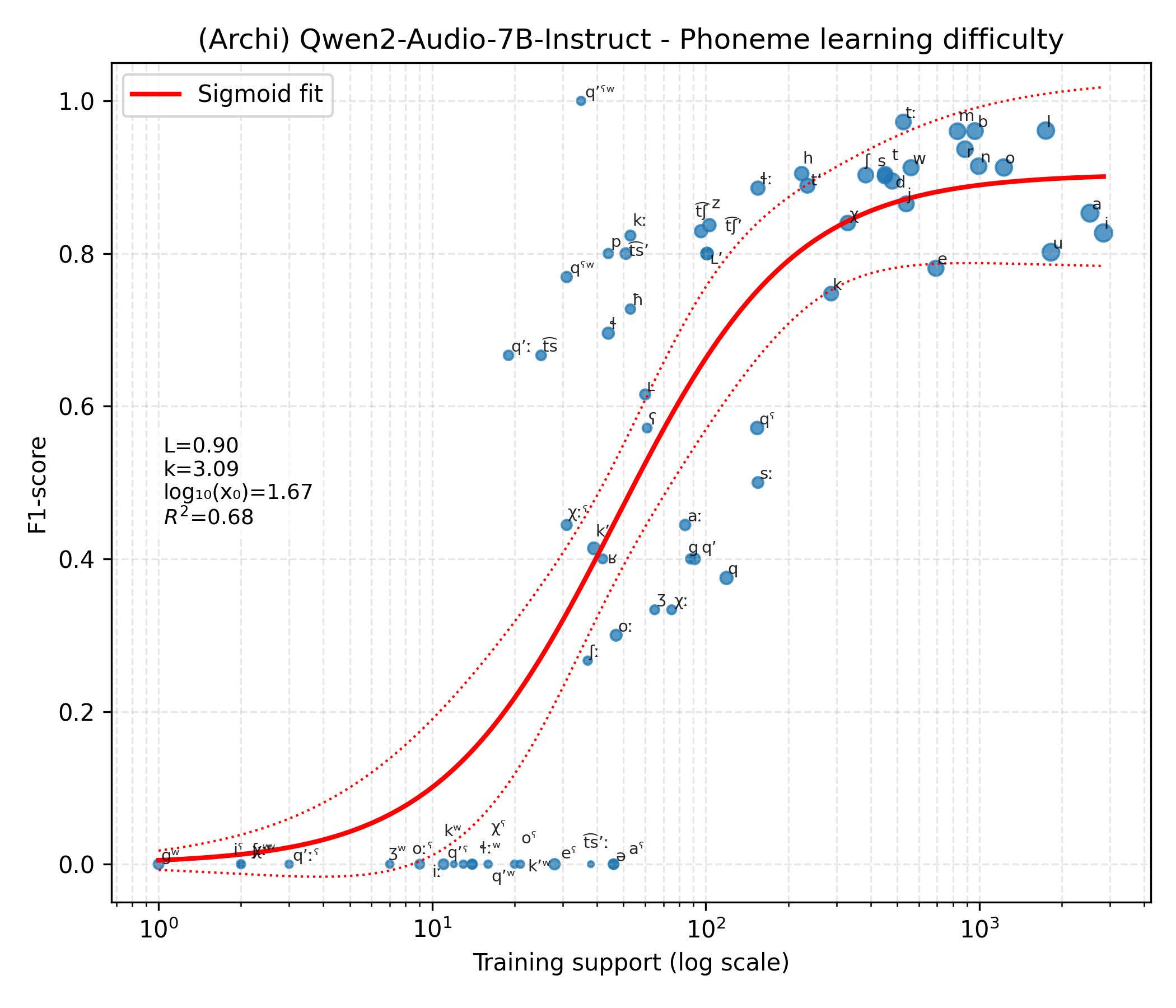}}
    \subfigure[]{\includegraphics[width=0.16\textwidth]{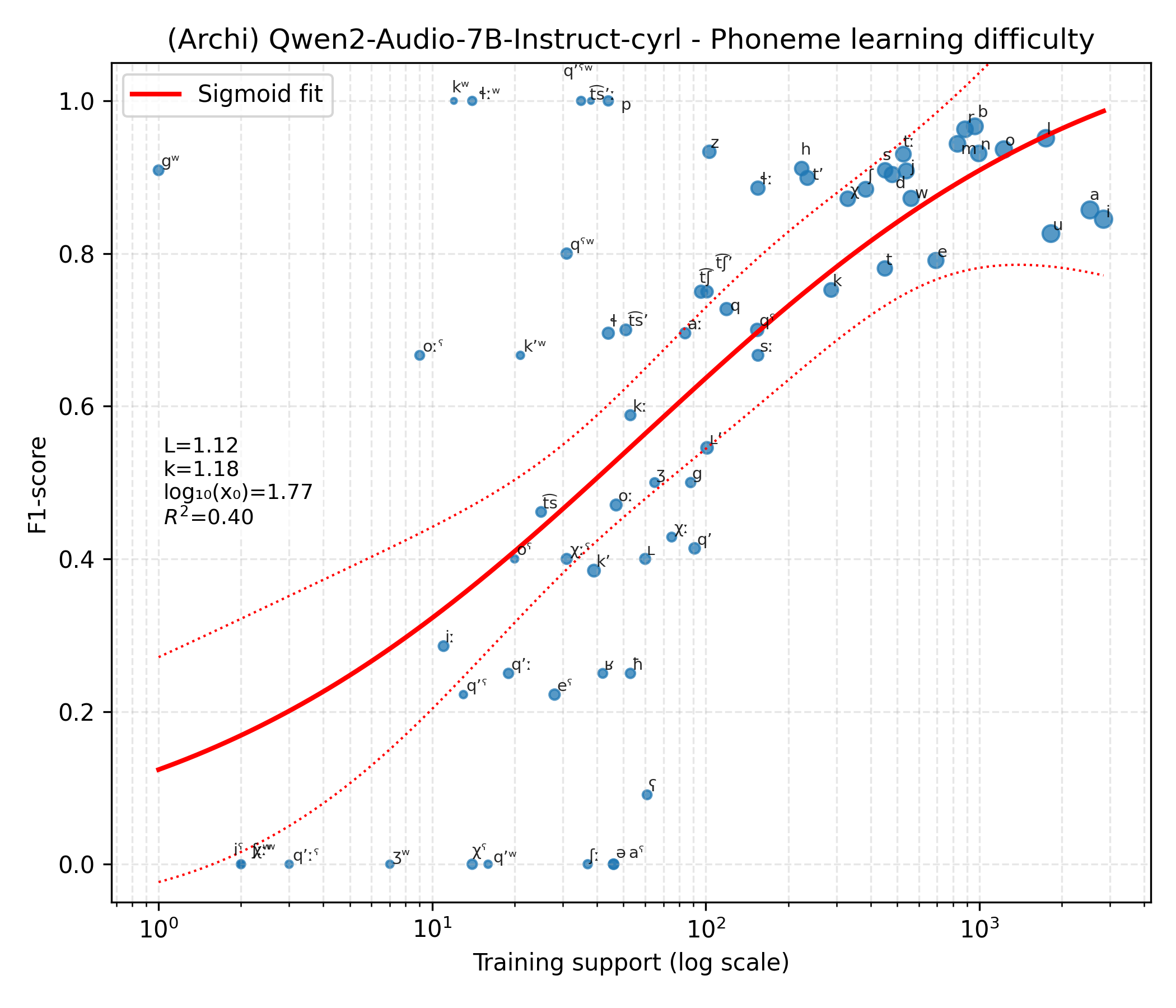}}
    \subfigure[]{\includegraphics[width=0.16\textwidth]{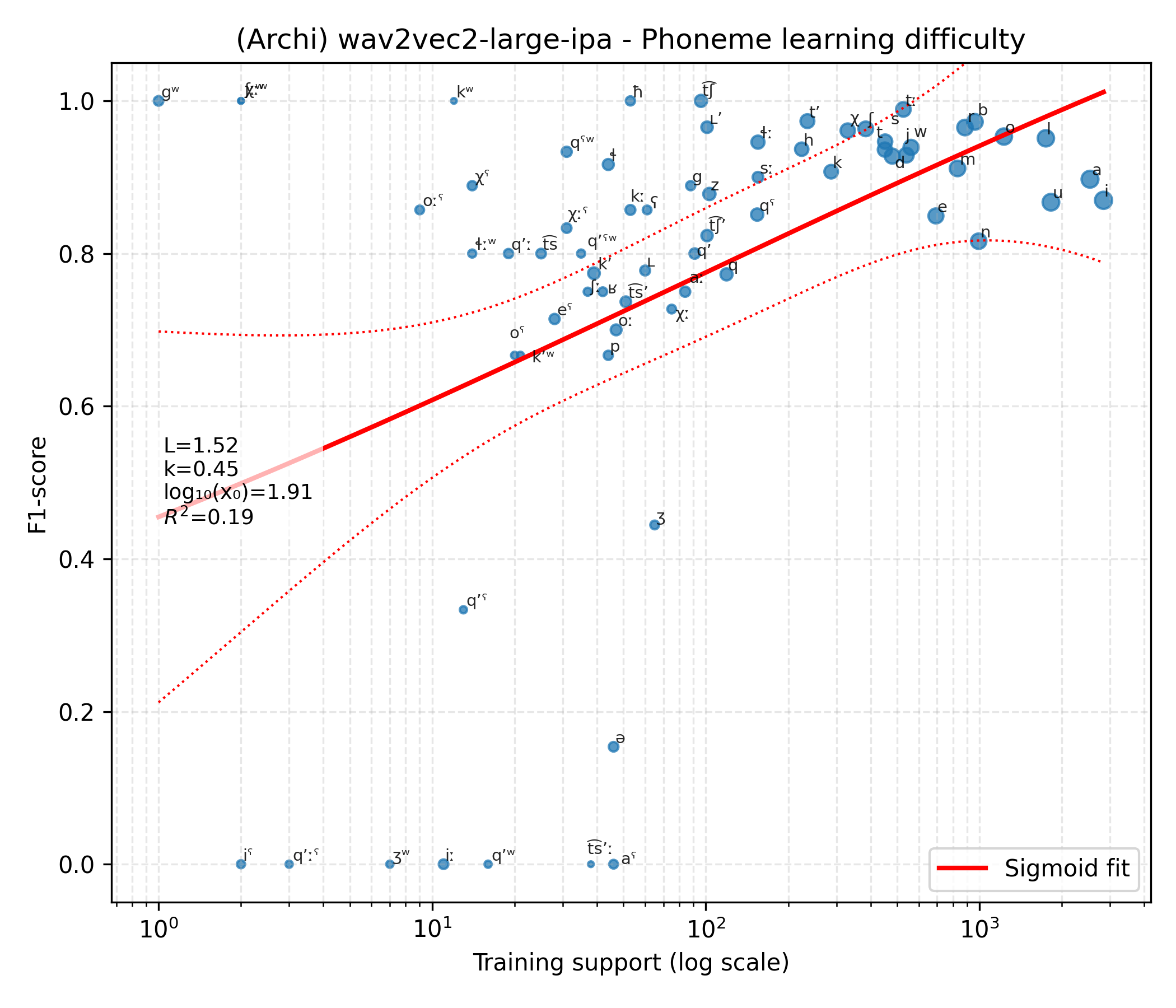}}
    \subfigure[]{\includegraphics[width=0.16\textwidth]{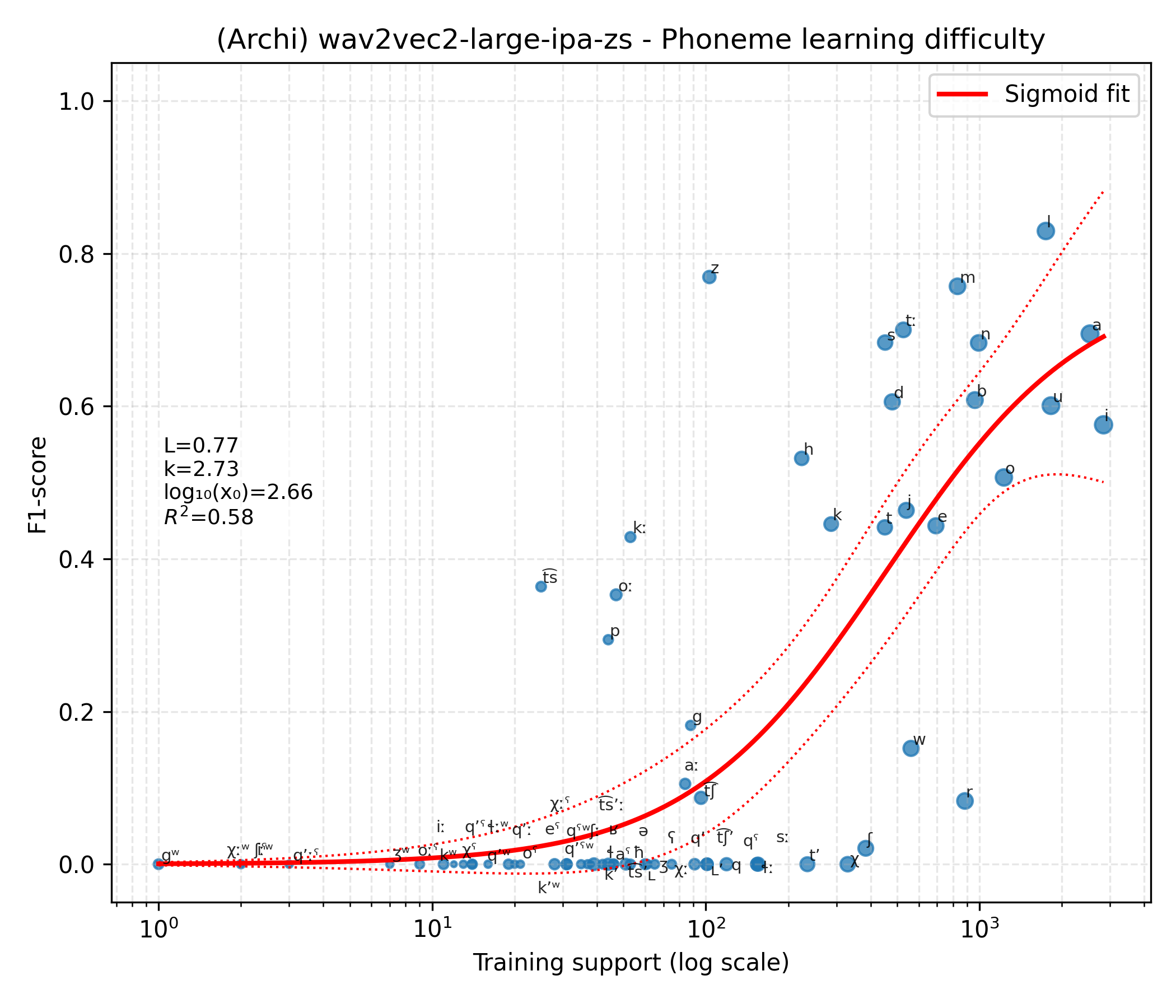}}
    \subfigure[]{\includegraphics[width=0.16\textwidth]{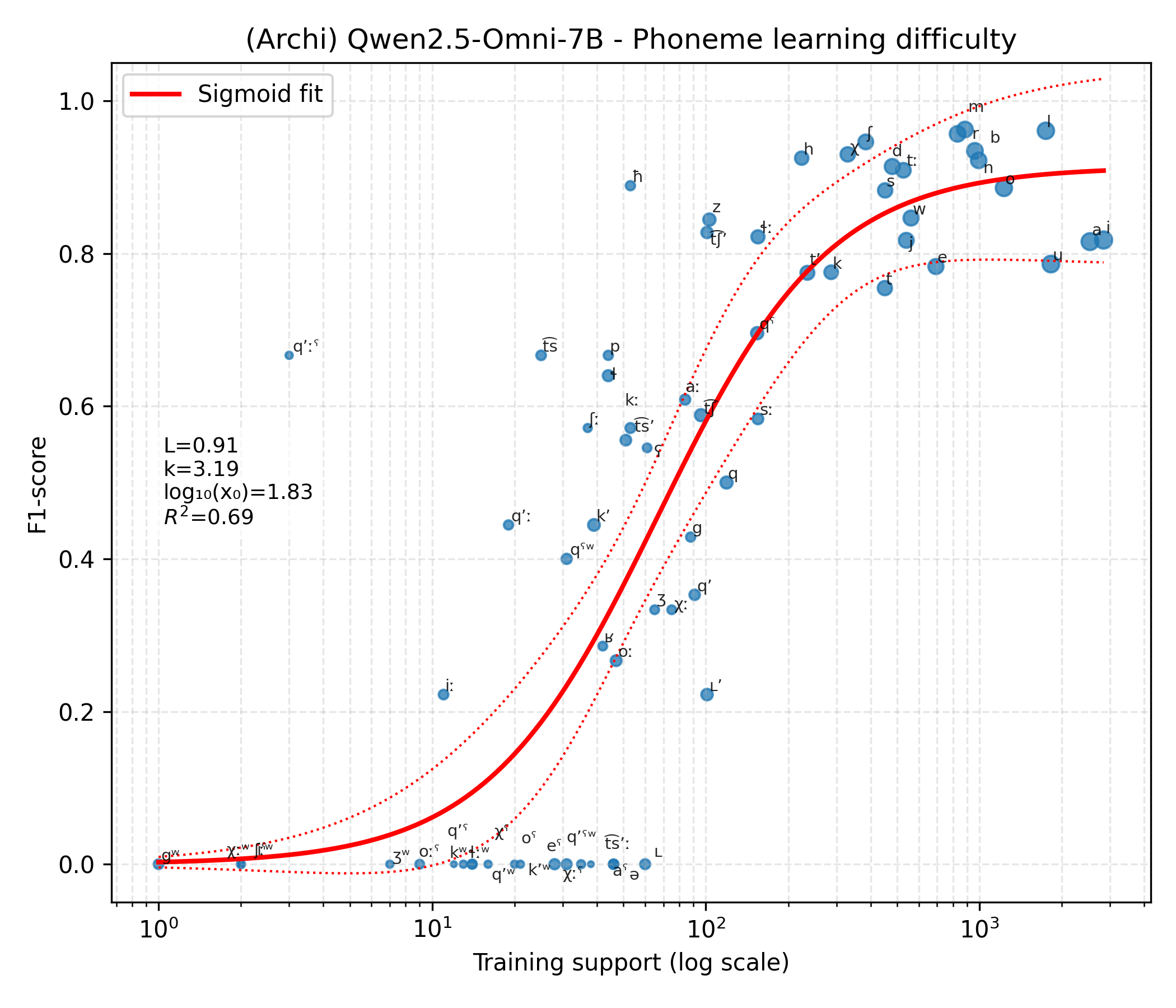}}
    \subfigure[]{\includegraphics[width=0.16\textwidth]{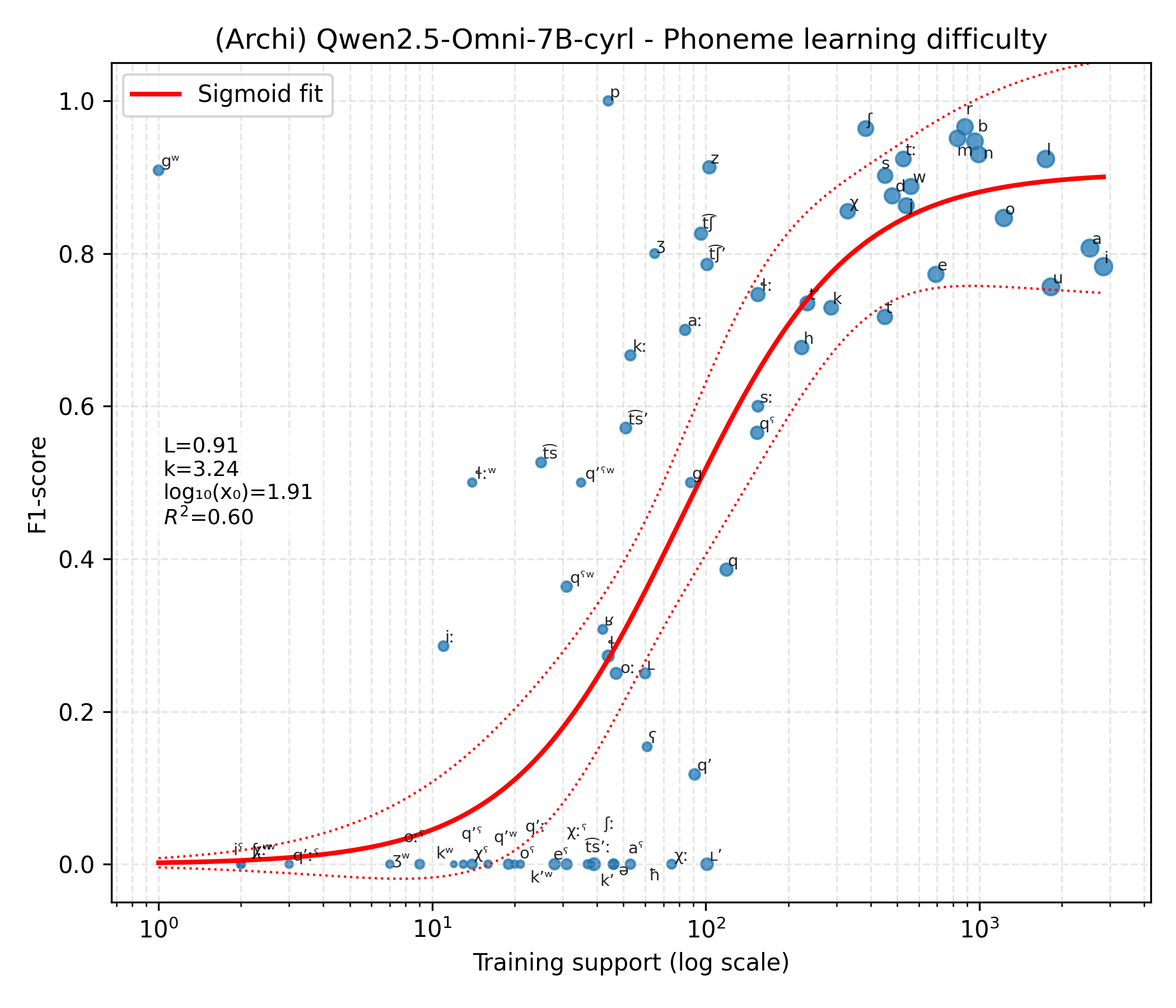}}
    \subfigure[]{\includegraphics[width=0.16\textwidth]{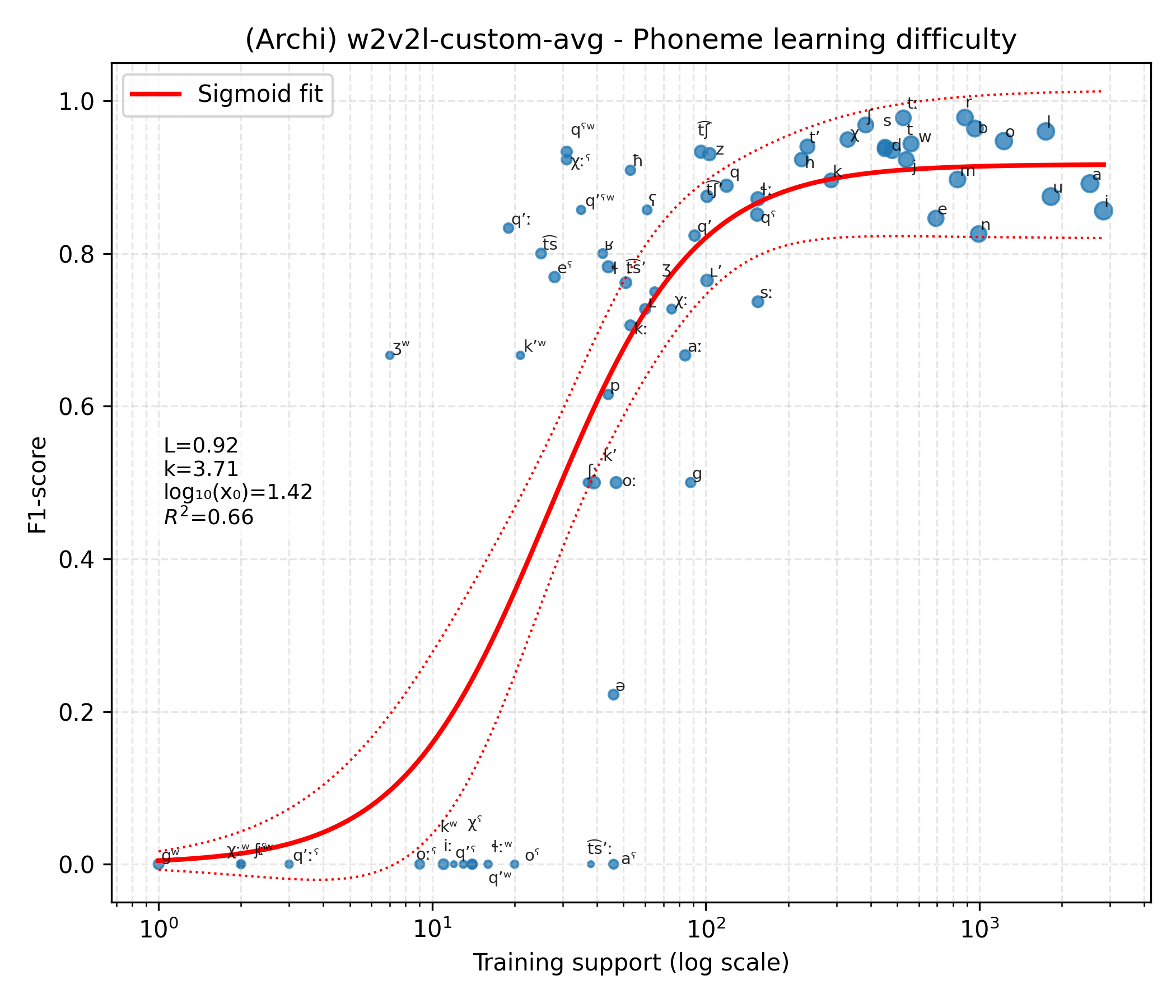}}
    \subfigure[]{\includegraphics[width=0.16\textwidth]{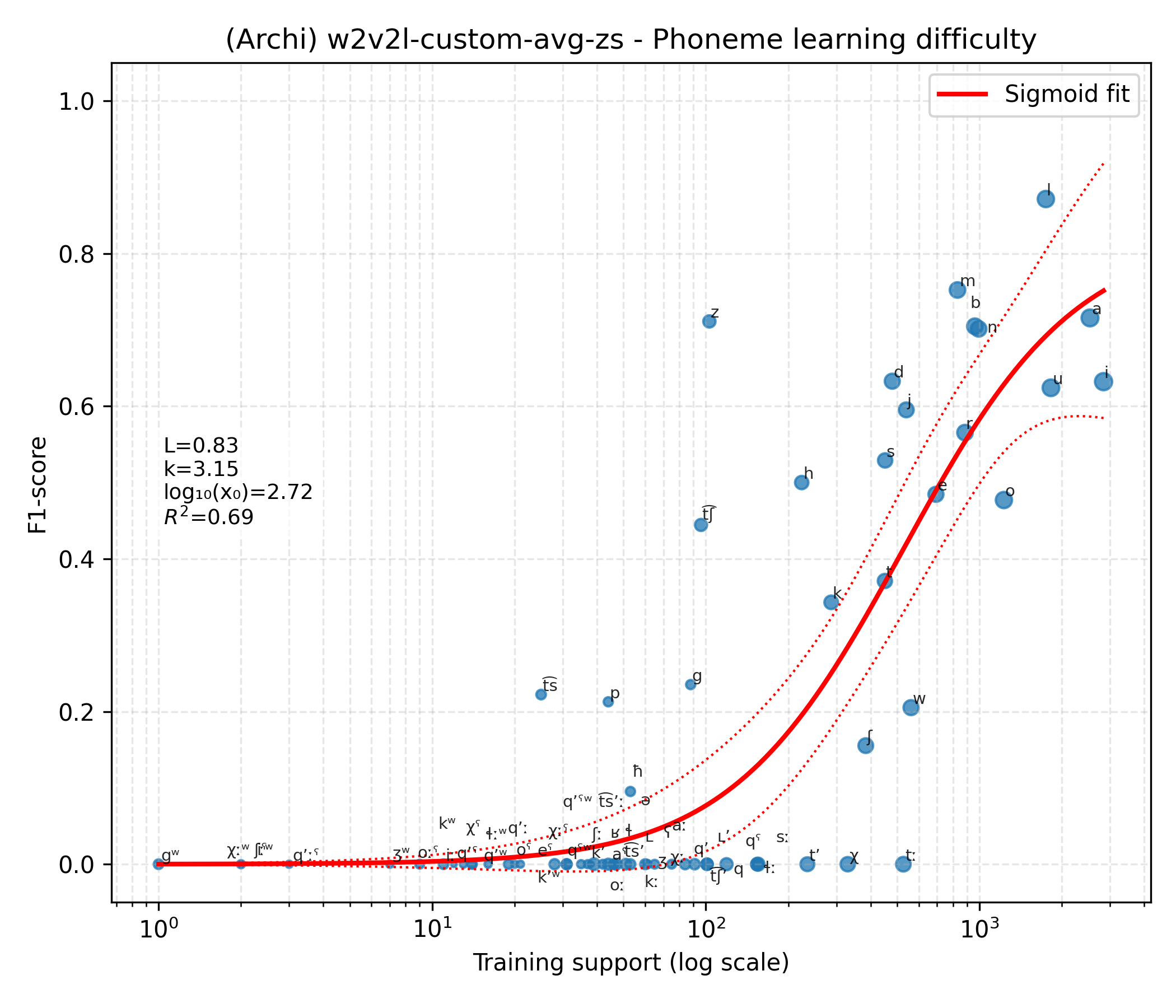}}
    \subfigure[]{\includegraphics[width=0.16\textwidth]{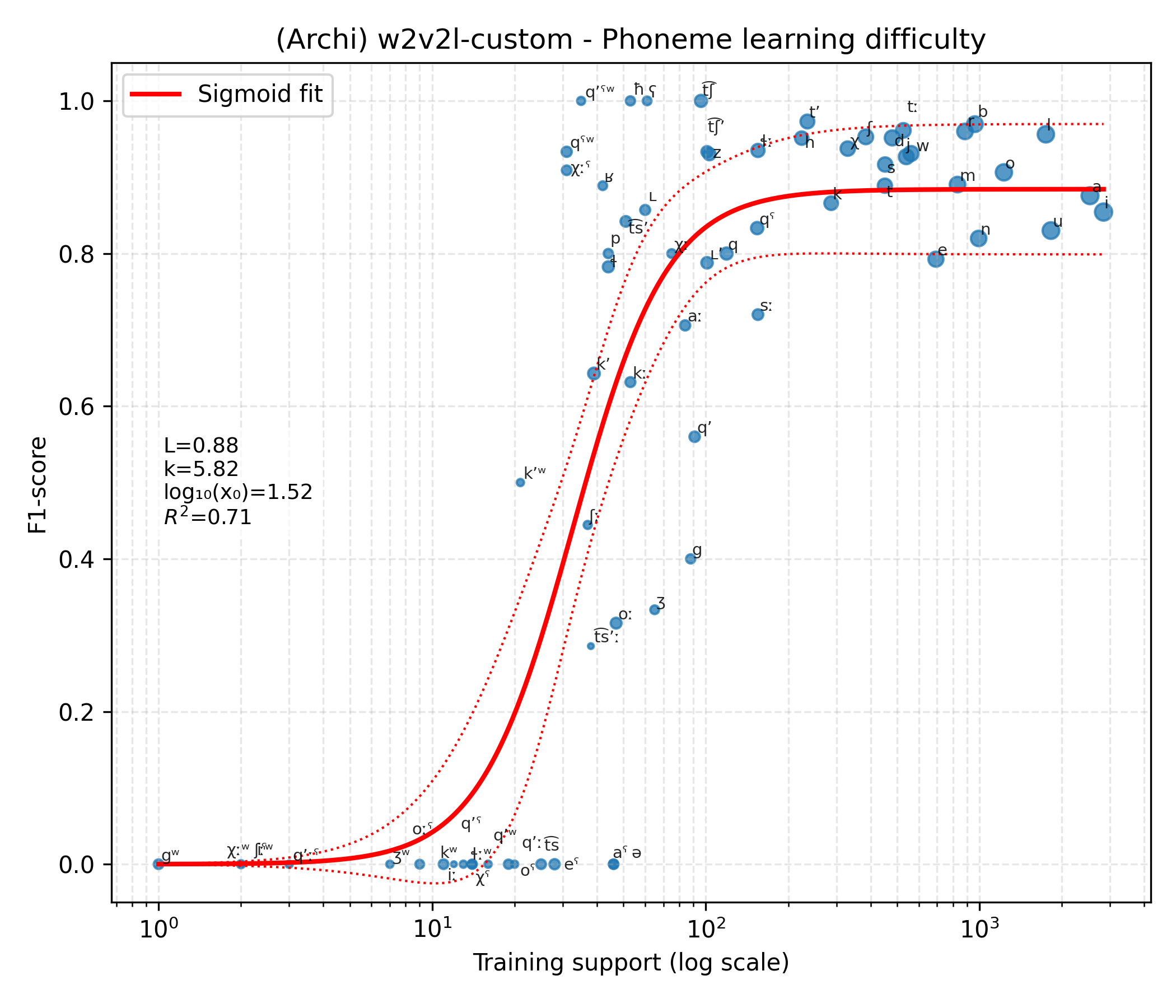}}
    \subfigure[]{\includegraphics[width=0.16\textwidth]{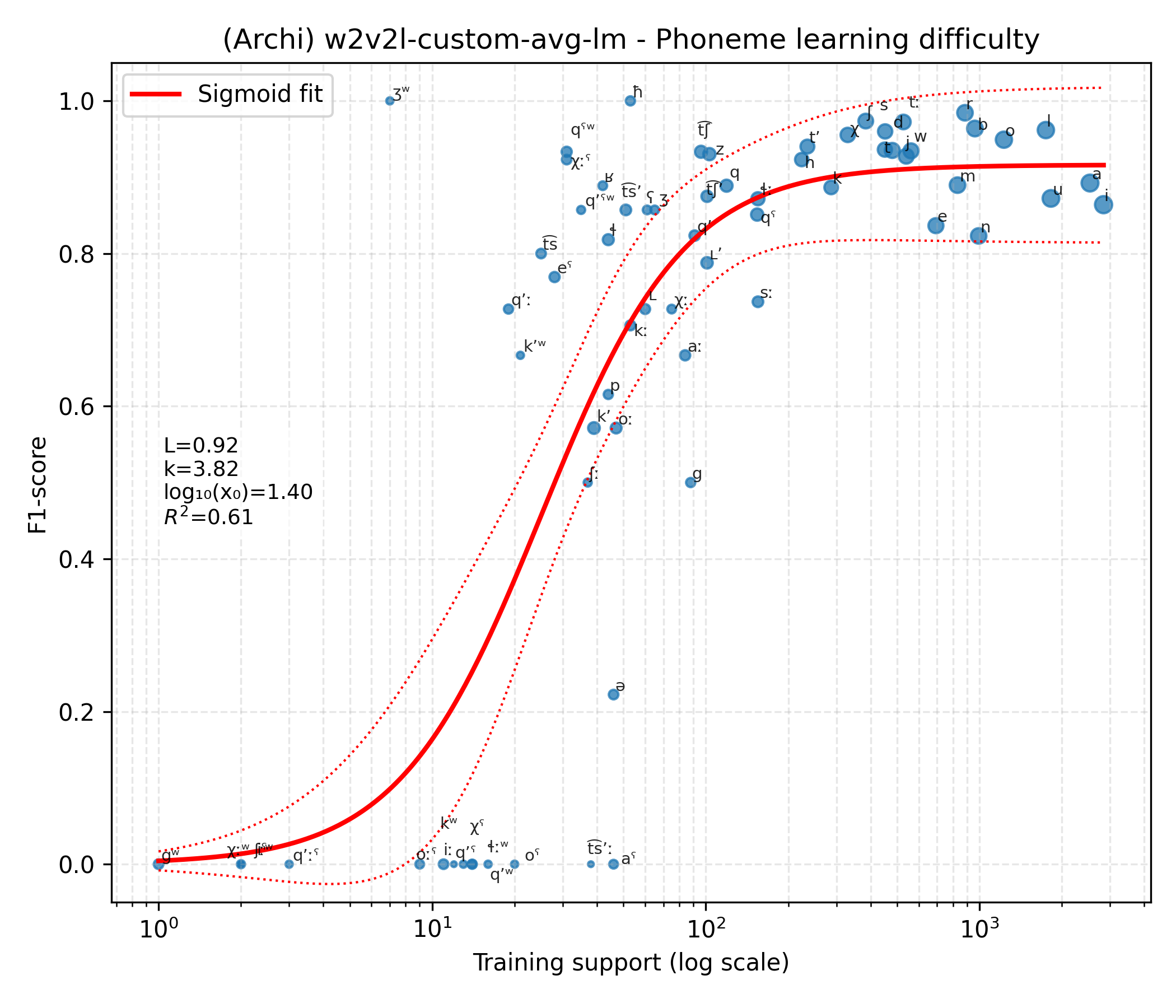}}
    \subfigure[]{\includegraphics[width=0.16\textwidth]{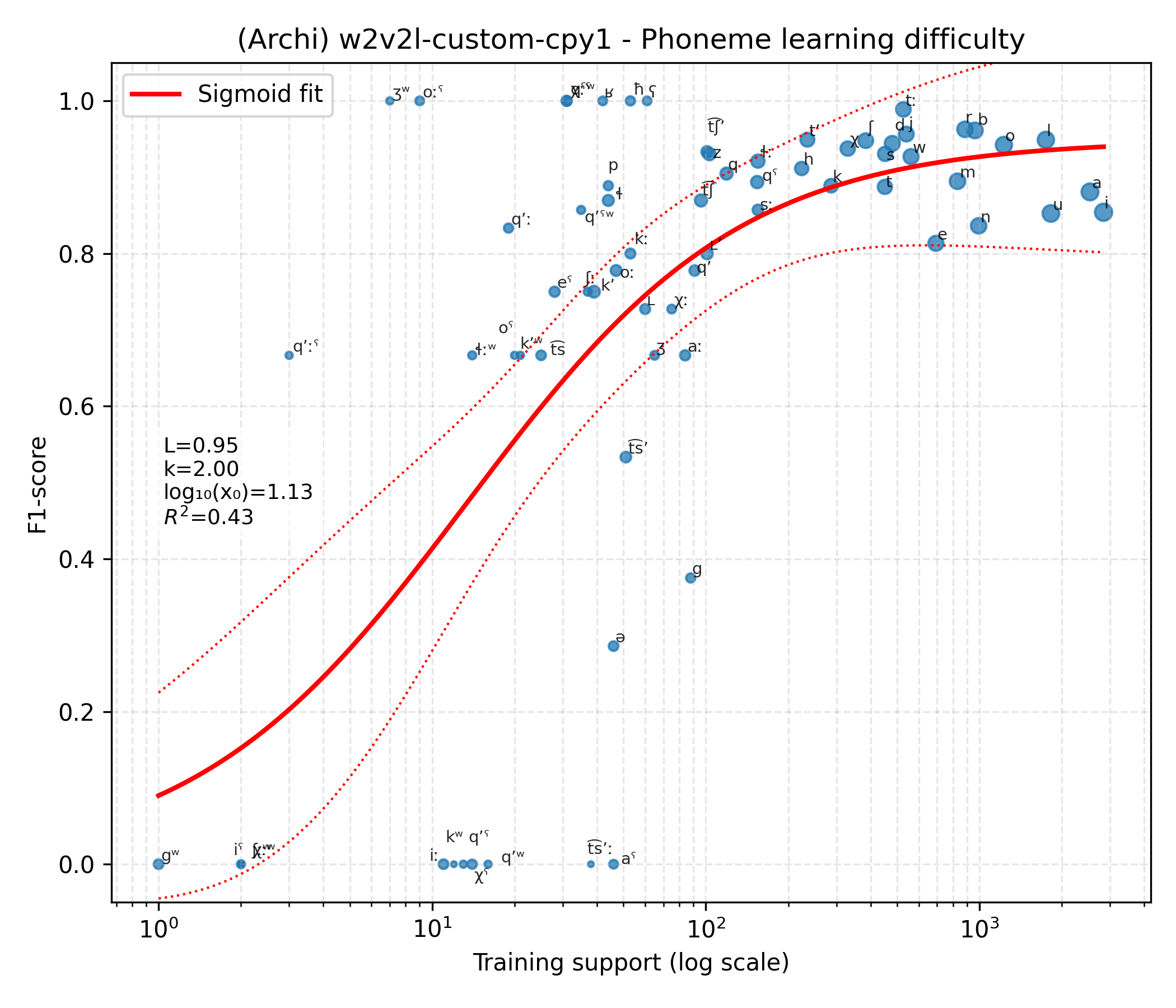}}
    \subfigure[]{\includegraphics[width=0.16\textwidth]{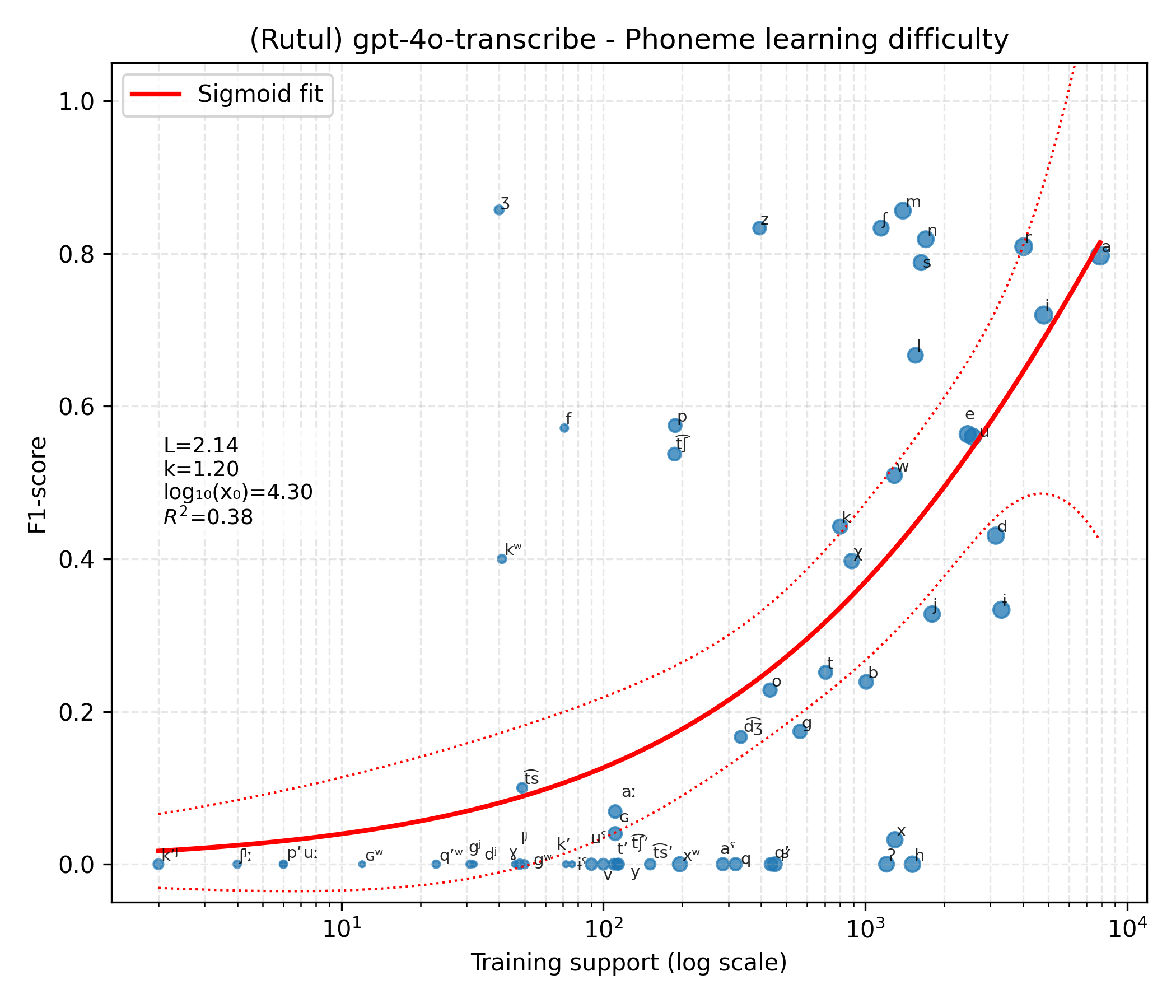}}
    \subfigure[]{\includegraphics[width=0.16\textwidth]{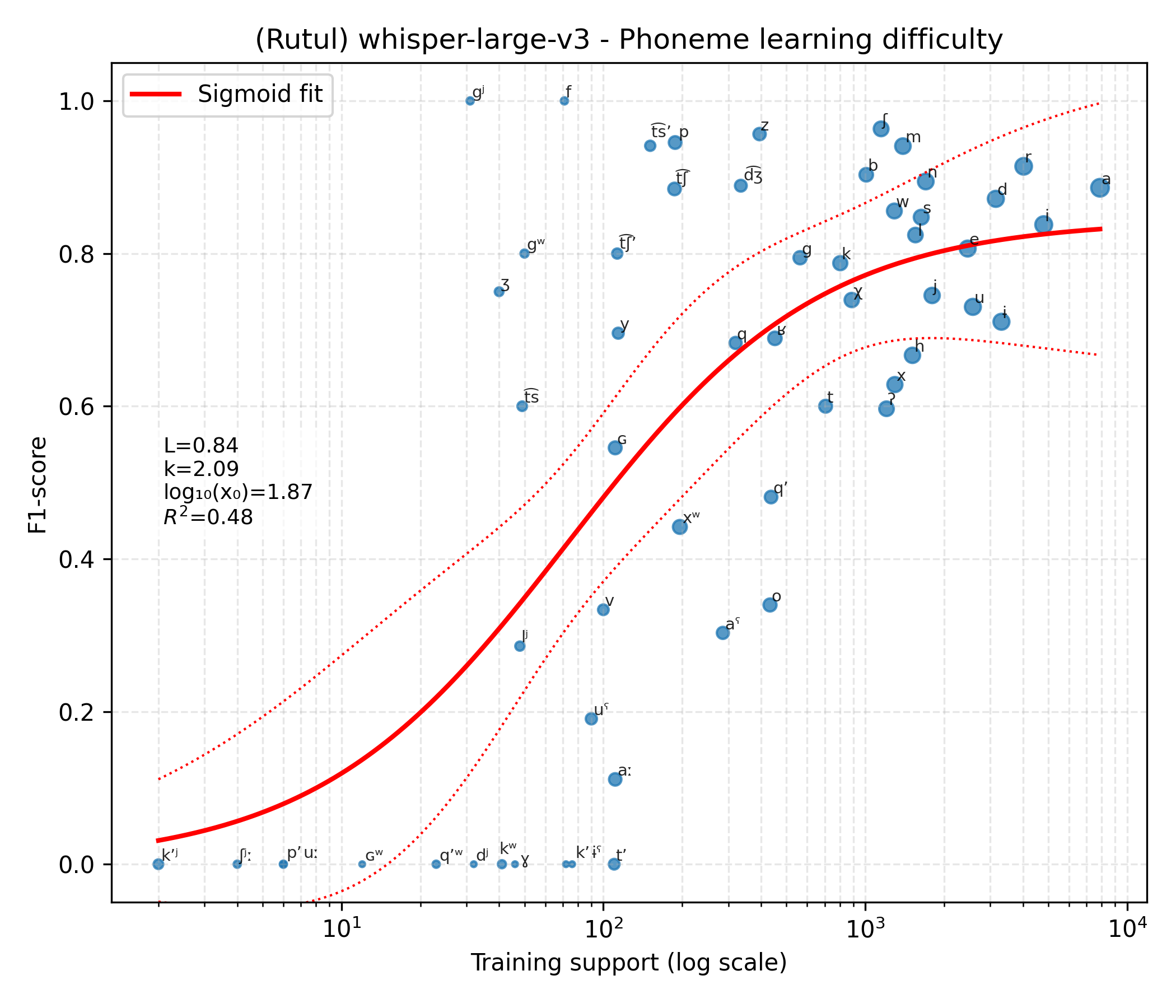}}
    \subfigure[]{\includegraphics[width=0.16\textwidth]{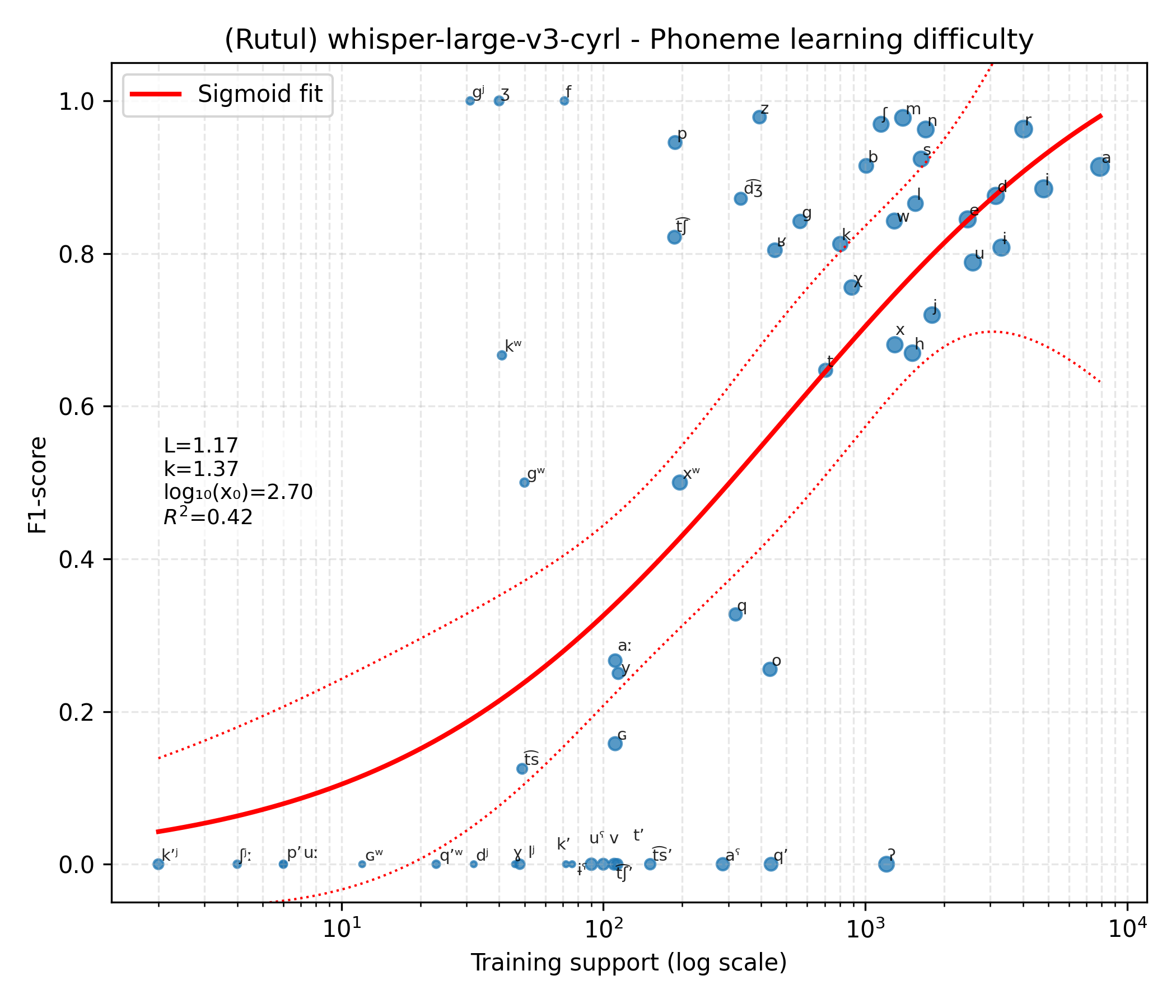}}
    \subfigure[]{\includegraphics[width=0.16\textwidth]{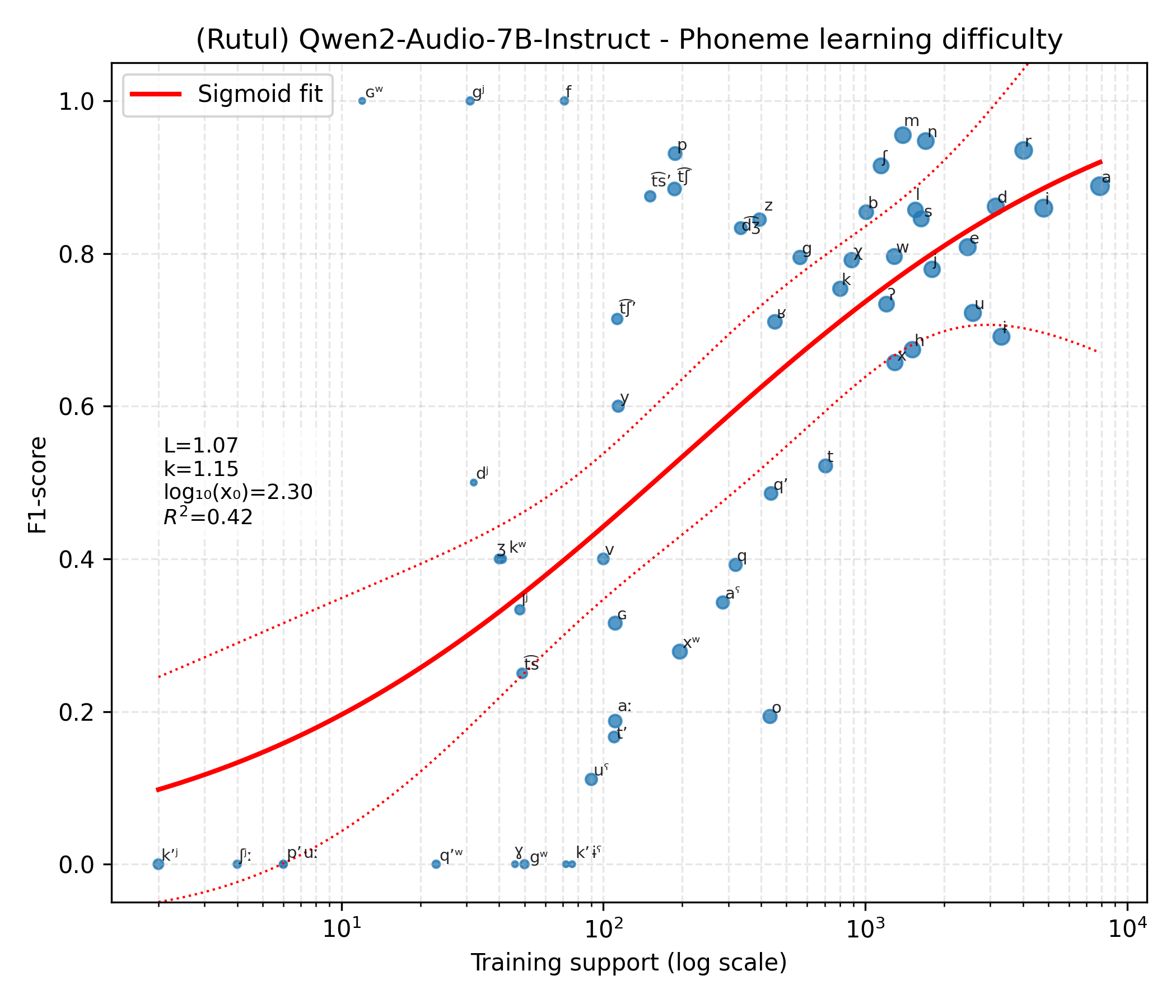}}
    \subfigure[]{\includegraphics[width=0.16\textwidth]{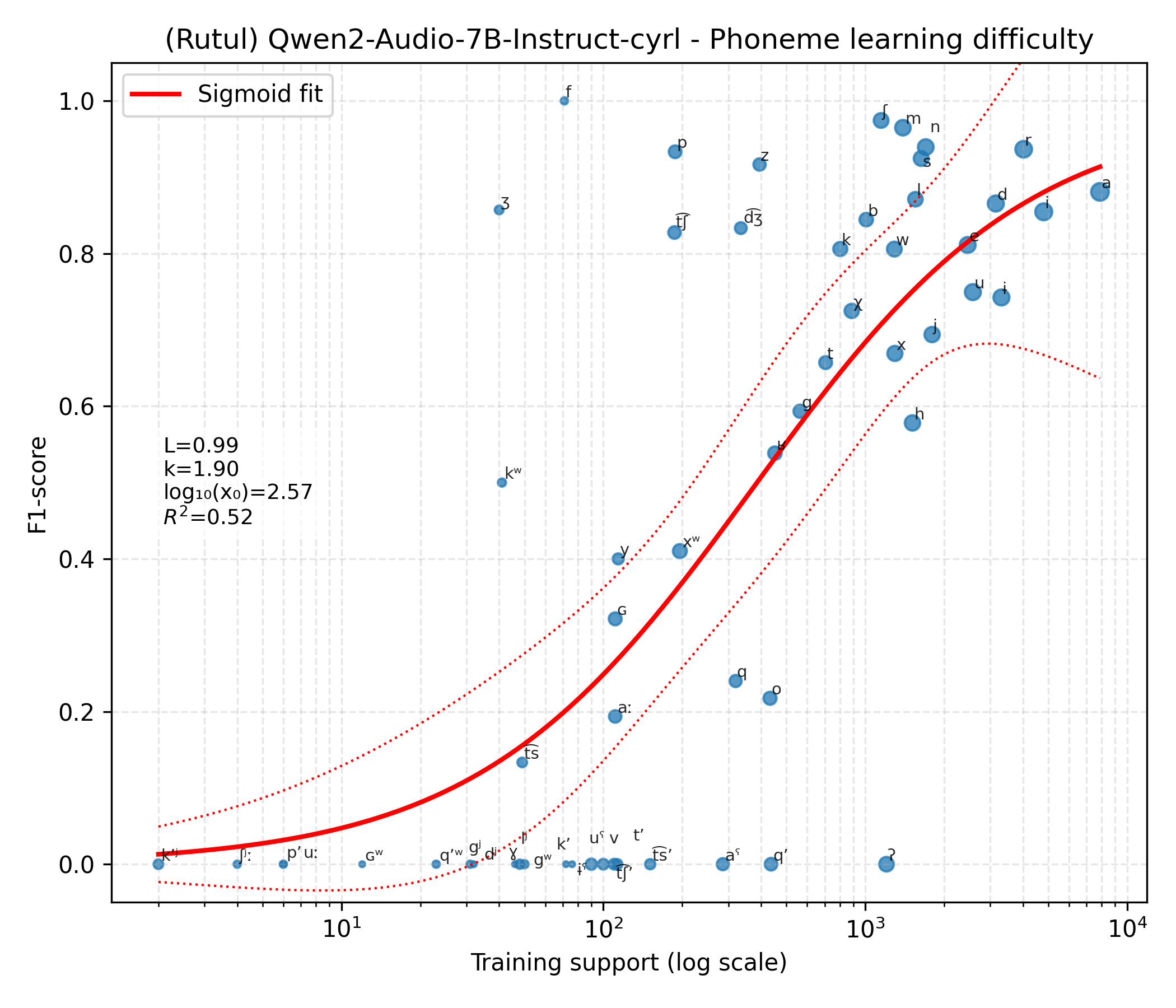}}
    \subfigure[]{\includegraphics[width=0.16\textwidth]{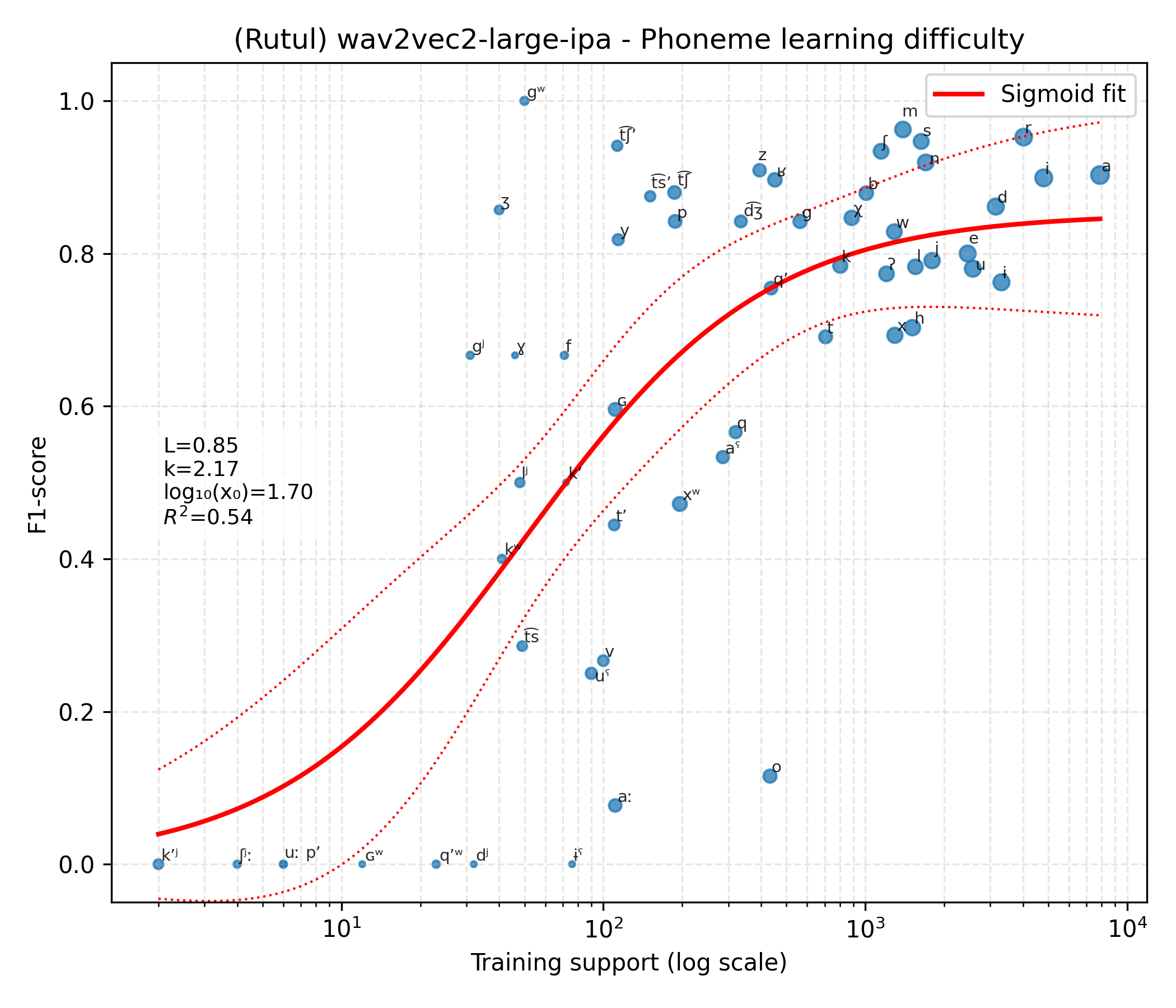}}
    \subfigure[]{\includegraphics[width=0.16\textwidth]{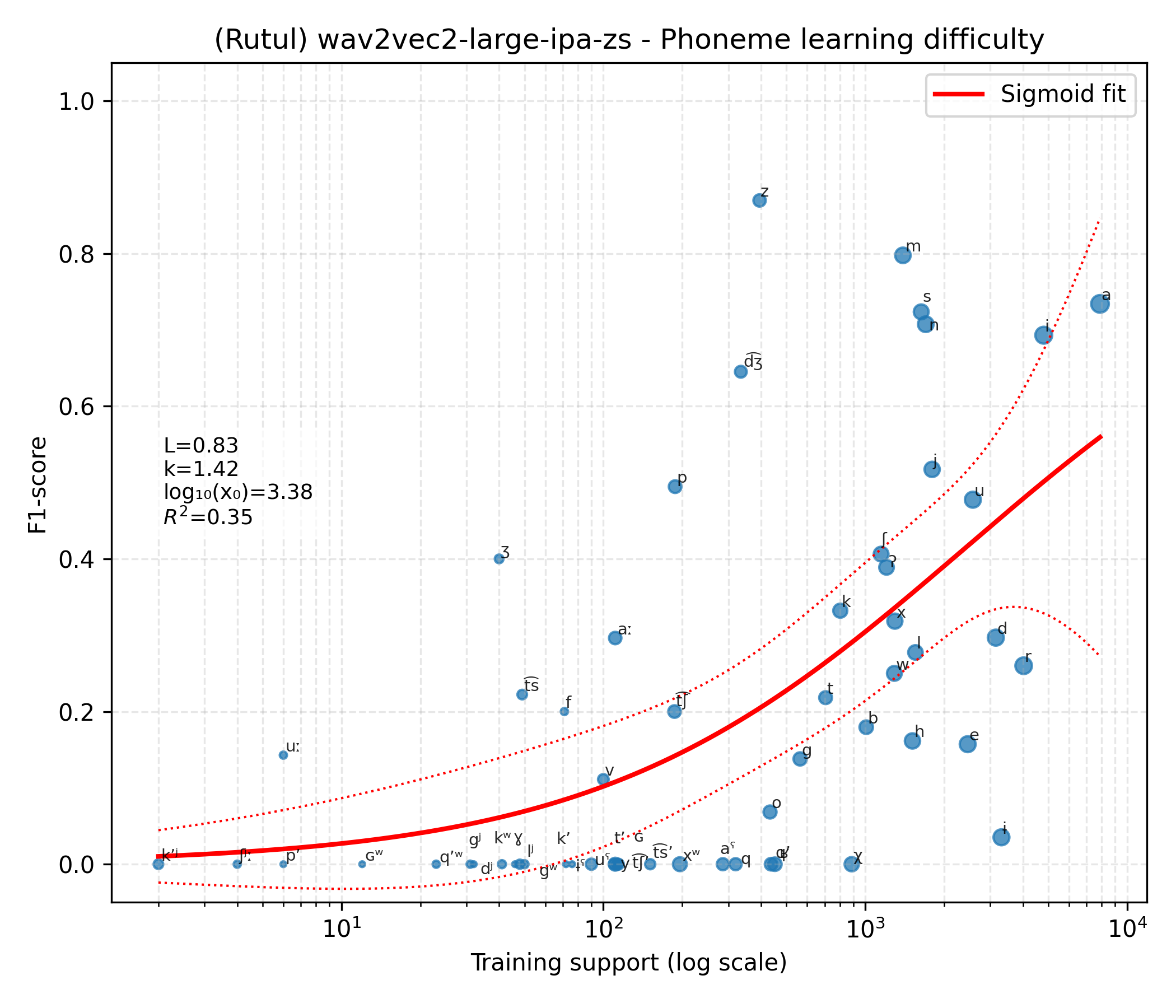}}
    \subfigure[]{\includegraphics[width=0.16\textwidth]{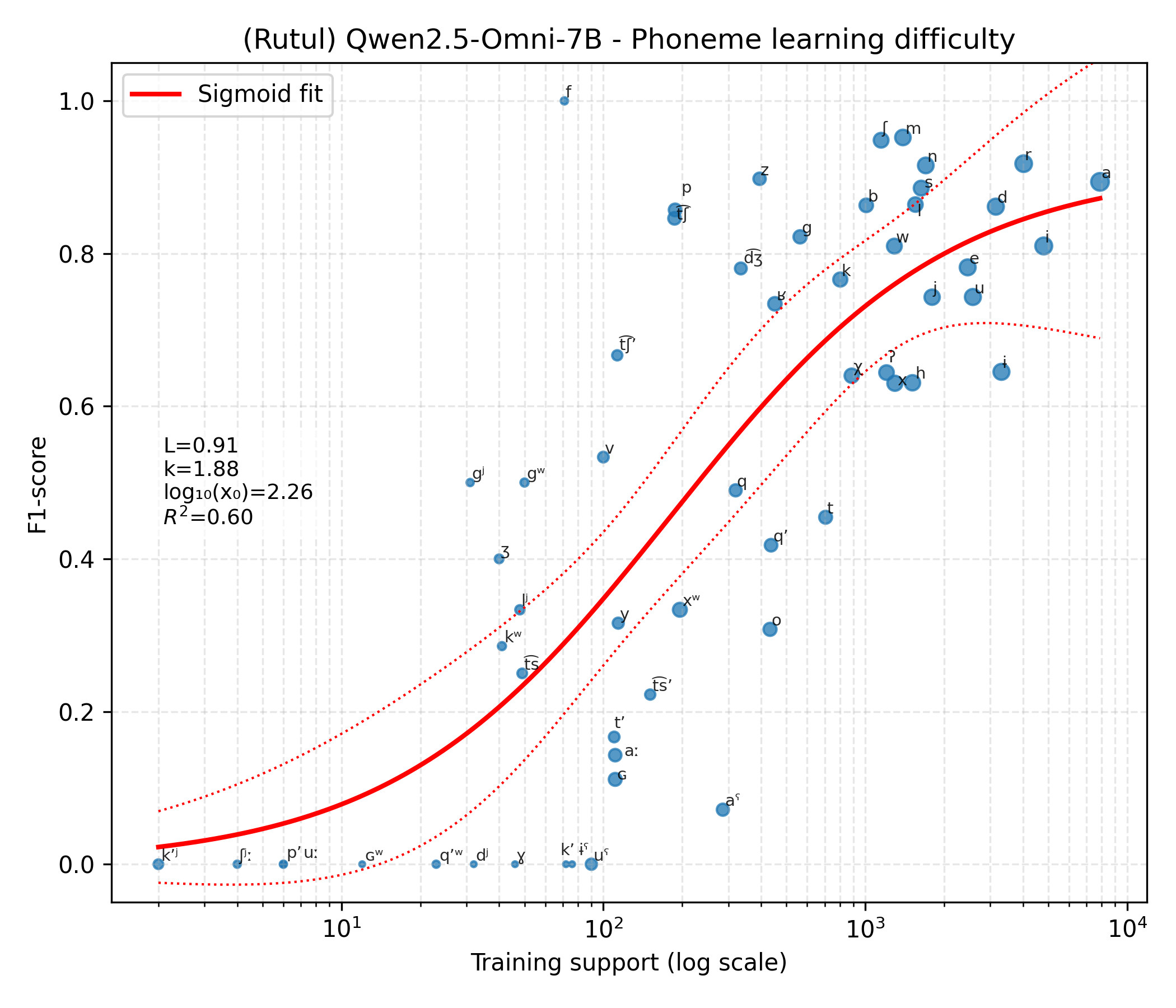}}
    \subfigure[]{\includegraphics[width=0.16\textwidth]{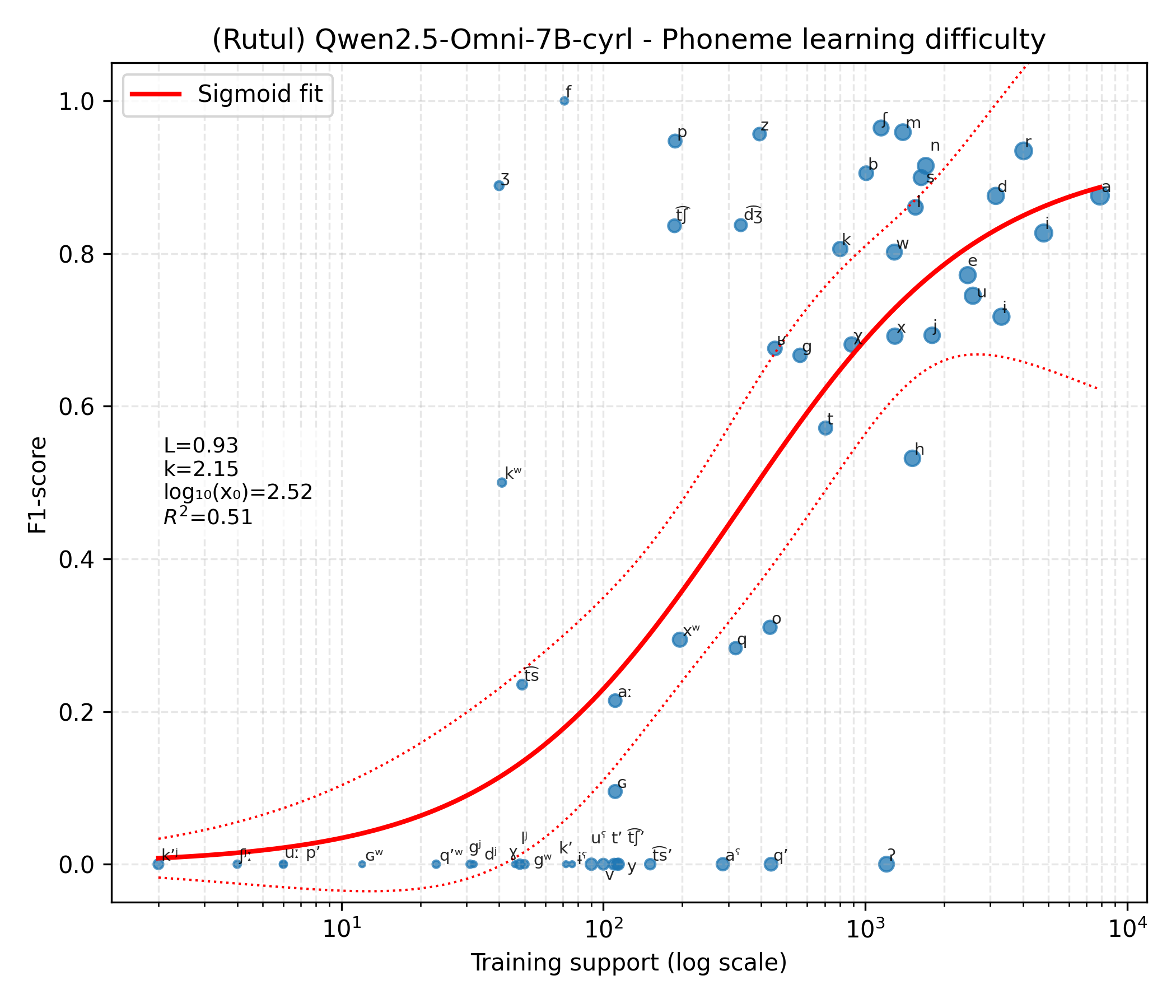}}
    \subfigure[]{\includegraphics[width=0.16\textwidth]{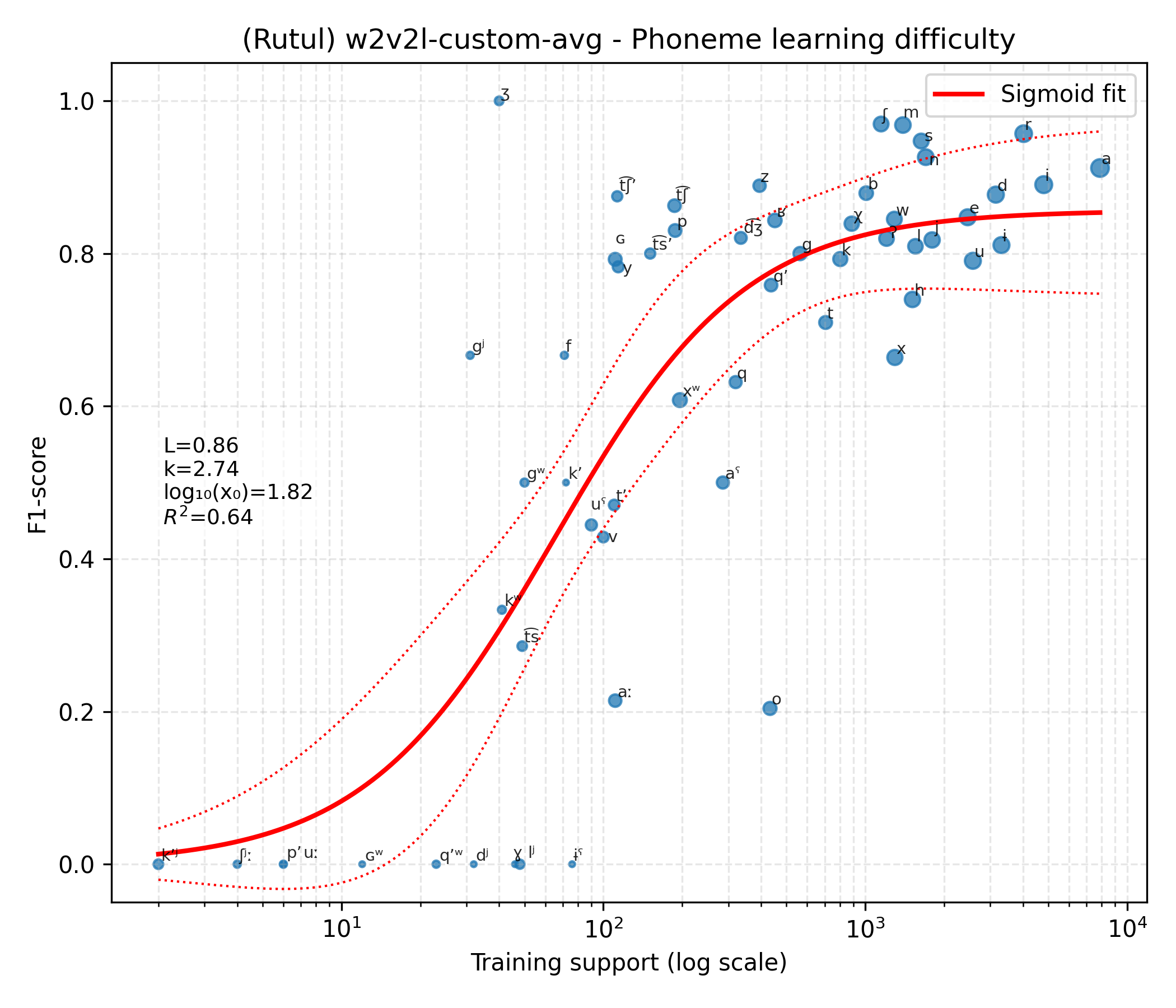}}
    \subfigure[]{\includegraphics[width=0.16\textwidth]{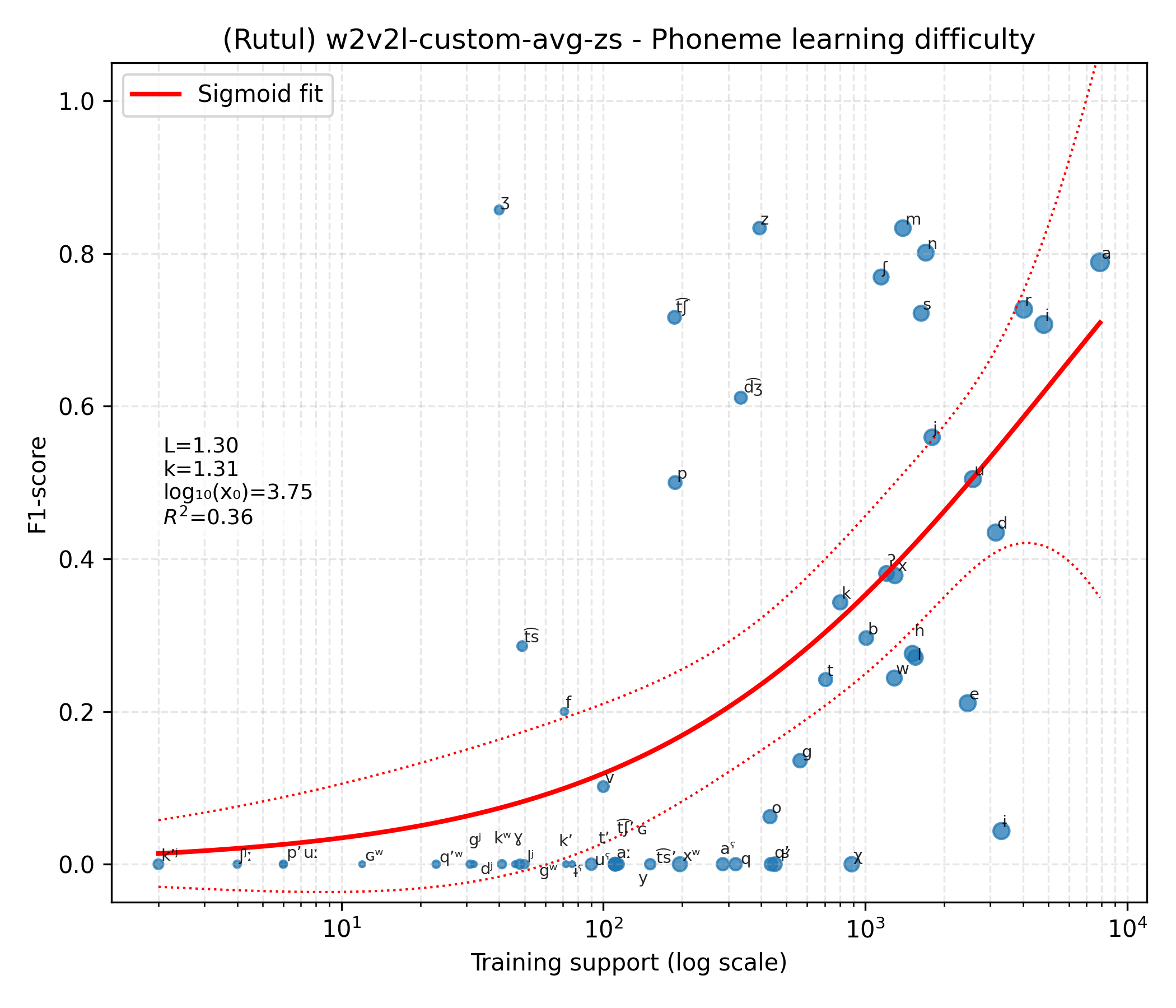}}
    \subfigure[]{\includegraphics[width=0.16\textwidth]{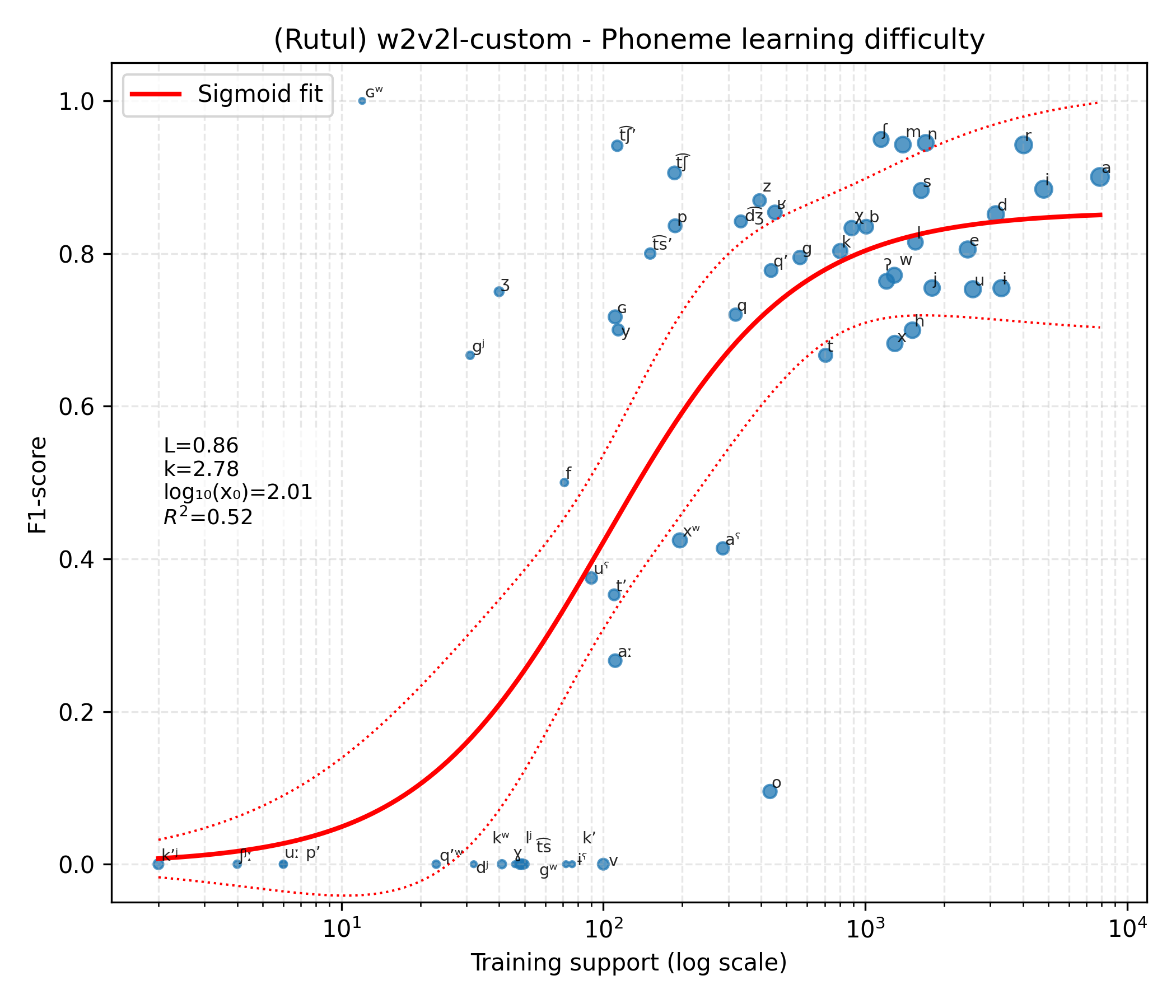}}
    \subfigure[]{\includegraphics[width=0.16\textwidth]{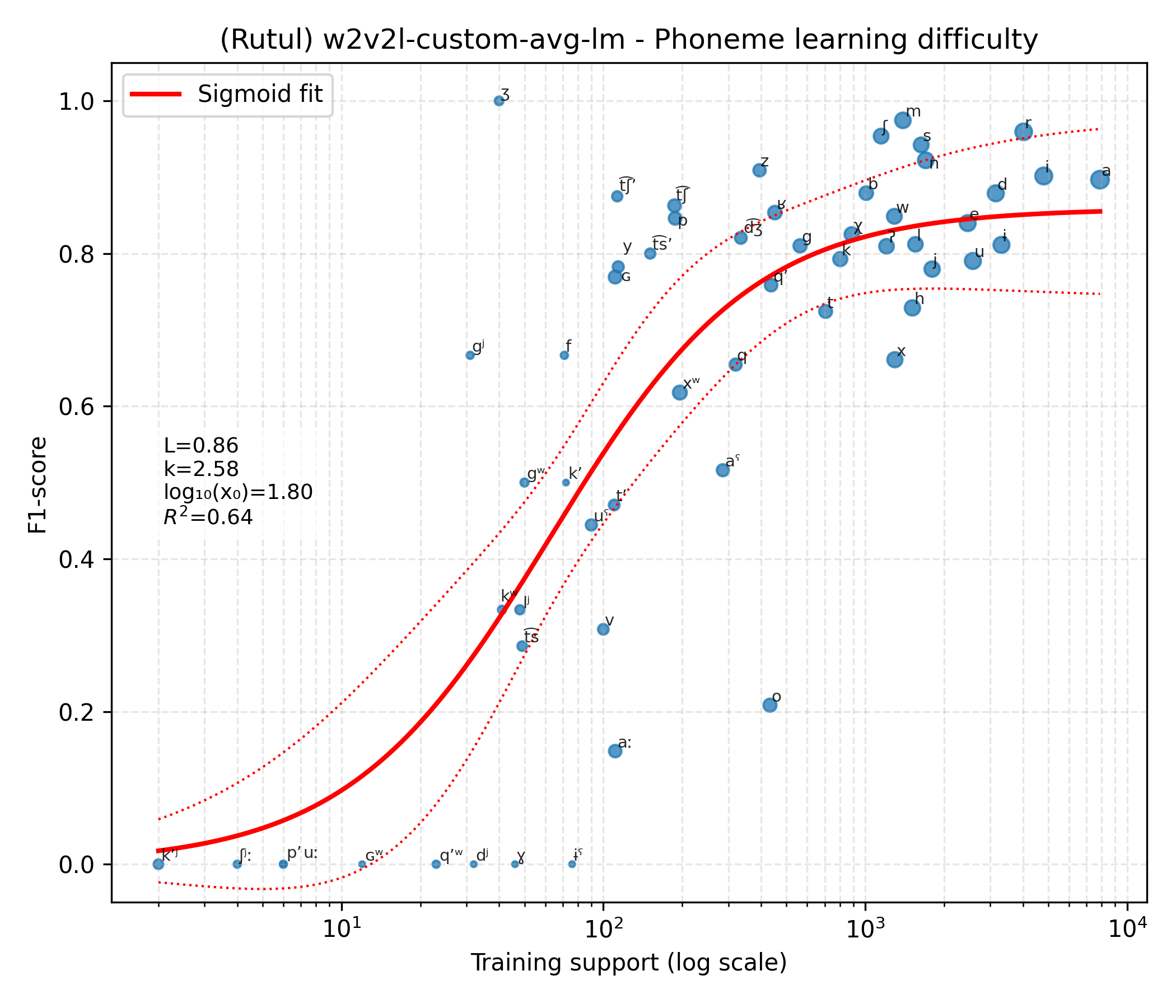}}
    \subfigure[]{\includegraphics[width=0.16\textwidth]{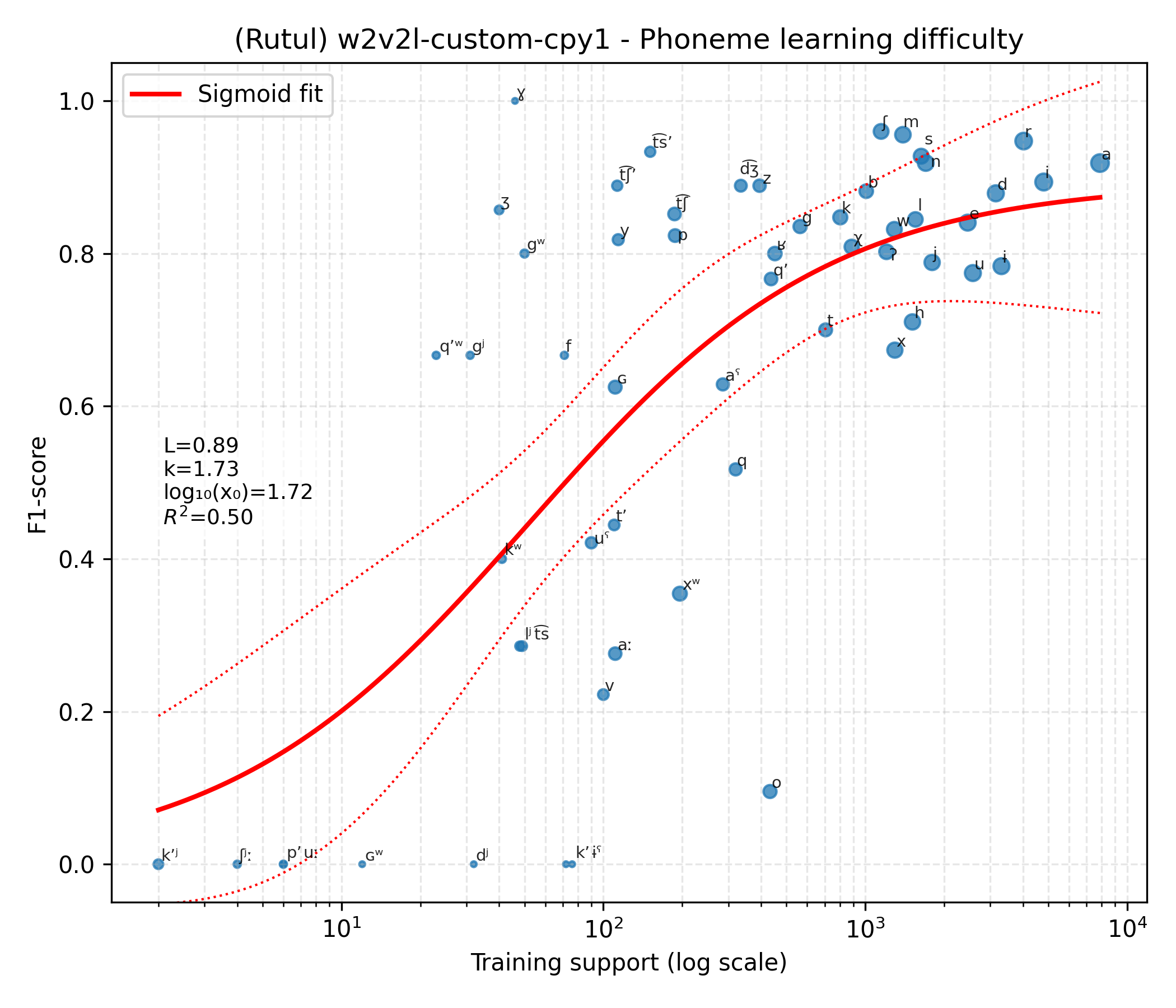}}
    \caption{Phoneme-level F1-scores against (log) training frequencies, illustrating a characteristic sigmoid-shaped learning trend in most cases. Exceptions are zero-shot applications (a, g, k, o, u, y) and cases demonstrating few-shot capabilities (b, c).}
    \label{fig:sigmoids}
\end{figure*}

\subsection{Training Frequency–F1 Score Analysis}

A central observation of this work is the strong relationship between phoneme-level F1 scores and the logarithm of training frequency across models and both languages (Figure~\ref{fig:sigmoids}). As formalized in \S\ref{subsec:exp-sigmoid}, we fit a logistic curve to this relationship. For most fine-tuned models, phoneme recognition accuracy follows a characteristic sigmoid shape: very rare phonemes exhibit near-zero F1, followed by a sharp transition as training support increases, and eventual saturation for frequent phonemes. The resulting fits explain a substantial fraction of phoneme-level variability, with $R^2$ values typically in the $0.45$--$0.70$ range.

This pattern implies that many phonemes traditionally described as ``complex’’---including labialized, ejective, and pharyngealized segments---are difficult for ASR systems primarily because they are rare in the training data. Figure~\ref{fig:intro-sigmoid} illustrates this effect for a representative model–language pair, where each point corresponds to a phoneme, the horizontal axis denotes log training frequency, and point size reflects log (test frequency+1). The fitted sigmoid captures the dominant trend, while deviations reveal model- and language-specific effects.

Notable deviations from the sigmoid trend occur for whisper-large-v3 and wav2vec2-large-ipa on Archi, where several extremely low-frequency phonemes achieve higher F1 scores than predicted by the fitted curve. These cases are characterized by weaker overall fits (lower $R^2$), suggesting few-shot transfer from multilingual pretraining. Importantly, this behavior is not observed for Rutul, where the more conversational and spontaneous speech style (while the Archi corpus is read out, see \S\ref{sec:datasets}) leads these models to adhere more closely to the average sigmoid behavior.

To further probe these cases, we analyze low-support phonemes ($< 10^{1.6}$) by matching them to high-support counterparts ($> 10^{1.9}$) sharing the same base segment. Many well-performing low-support phonemes (F1 $>$ 0.8) are systematic diacritic variants of frequent bases (e.g., a\textipa{:}/a\textipa{\super Q} $\rightarrow$ a; e\textipa{:}/e\textipa{\super Q} $\rightarrow$ e; o\textipa{:}/o\textipa{\super Q} $\rightarrow$ o; \textipa{S:}/\textipa{S:}{\super w} $\rightarrow$ \textipa{S}), with moderate correlation ($\rho \sim  0.4 - 0.49$). However, some consonants (e.g., g{\super w}, k{\super w}, \textipa{\t{tS}}') are not captured by this matching, suggesting that similarity-based effects may partly explain the deviations, while transfer from pretrained representations remains a plausible but inconclusive factor.

Zero-shot systems---wav2vec2-large-ipa-zs, w2v2l-custom-avg-zs, and gpt-4o-transcribe---also exhibit sigmoid-shaped trends, but with transition regions shifted toward higher training frequencies (larger $\log x_0$). As a result, even moderately frequent phonemes fall within the low-accuracy regime, highlighting that language-specific fine-tuning primarily acts to shift the sigmoid leftward, enabling effective learning at substantially lower levels of phoneme support.

Consistent with this interpretation, the estimated sigmoid midpoints ($\log x_0$) cluster around $1.6 \pm 0.3$ for Archi and $2.1 \pm 0.4$ for Rutul, excluding the anomalous cases. This suggests that achieving on the order of $10^2$ training instances per phoneme may be sufficient to move most segments into the steep learning regime, providing a practical target for future data collection efforts. The right-shift in Rutul’s estimated midpoint may partly reflect differences in speech conditions, as it consists of spontaneous speech recorded in noisier settings with greater speaker variability compared to the read speech in Archi (see \S\ref{sec:datasets}).

Overall, the frequency–F1 sigmoid provides a unifying explanation for phoneme-level error patterns across models.

\begin{figure}[t]
    \centering
    \subfigure[]{\includegraphics[width=0.7\columnwidth]{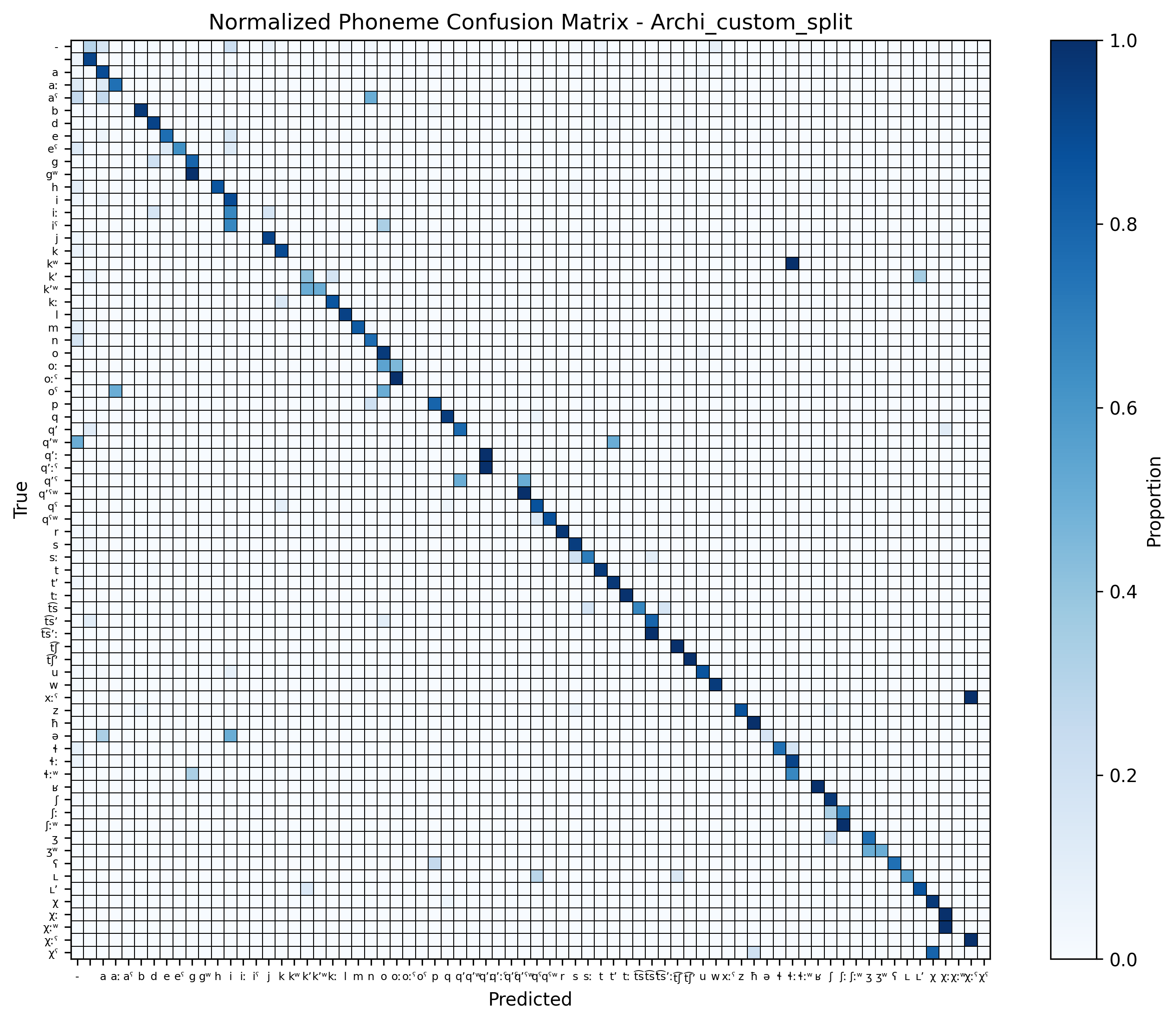}}
    \subfigure[]{\includegraphics[width=0.7\columnwidth]{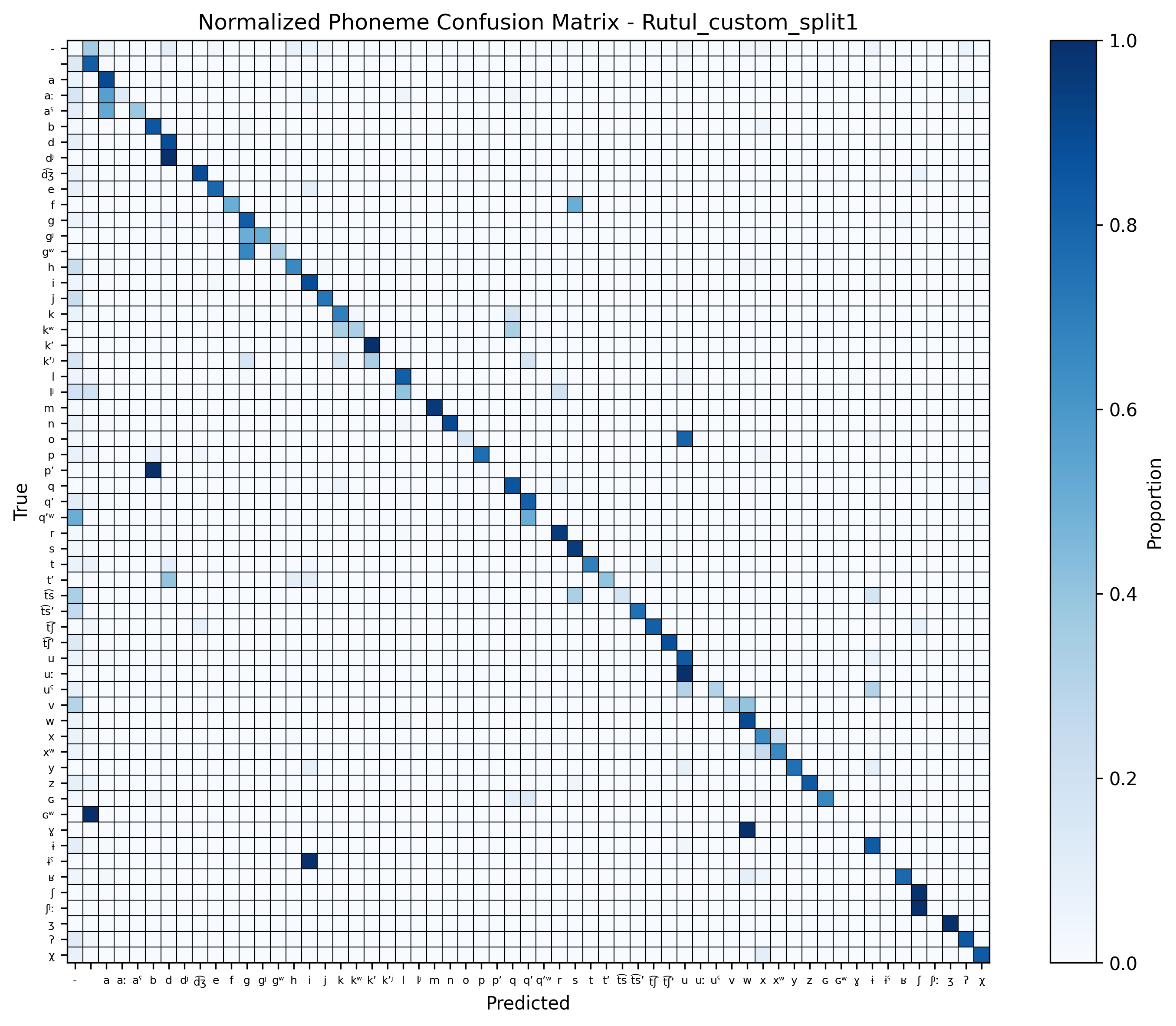}}
    \caption{Phoneme confusion matrices of model w2v2l-custom-avg on (a) Archi and (b) Kina Rutul}
    \label{fig:confmat}
\end{figure}

\subsection{Qualitative Error Analysis}

Figure~\ref{fig:confmat} shows normalized phoneme confusion matrices for the w2v2l-custom-avg model on Archi and Kina Rutul. Both matrices are strongly diagonal, indicating that most phonemes are correctly recognized, with errors concentrated in a small number of systematic confusions.

A recurring pattern in both languages is the reduction of marked phonemes to their unmarked counterparts. For example, long and pharyngealized vowels such as /i\textipa{:}/ and /i\textipa{\super Q}/ are frequently confused with the short vowel /i/. Similar effects are observed for consonants with secondary articulations, where labialized or pharyngealized segments are often mapped to their plain counterparts.

Other frequent error types include vowel quality confusions and incorrect word boundary detection. Examples (1) from Archi and (2) from Kina Rutul compare expert transcriptions with ASR outputs from the best-performing models (whisper-large-v3 for Archi and w2v2l-custom-avg for Kina Rutul).

\paragraph{(1) Archi example}

~\\\textit{Expert:} os \textipa{\t{tS}}'emna os bo\textipa{S}orm\b{i}n ha\b{lm}a\textipa{X}d\b{u} \b{e}wdili jat\textipa{:}i\b{k} {\scriptsize\;L}'arak war{\scriptsize\;L}irt\textipa{:}u i\b{k'\super w} \b{w}i\textipa{X}du hibat\textipa{:}u
\\\textit{ASR:} os \textipa{\t{tS}}'emna os bo\textipa{S}orm\b{u}n ha\b{l m}a\textipa{X}d \b{i}wdili jat\textipa{:}i\b{ } {\scriptsize\;L}'arak war{\scriptsize\;L}irt\textipa{:}u i\b{k} \b{u}i\textipa{X}du hibat\textipa{:}u 
\\\textit{Translation:} ``There was once a man who had a friend, one who would go all the way down for you, with a trustful heart, a good one.''

This example illustrates a case of incorrect word segmentation (\textit{halma\textipa{X}du}), attributable to ASR. Likewise, the output \textit{ui\textipa{X}du} instead of \textit{wi\textipa{X}du}, and the realization of /k'\super w/ as /k/, plausibly reflect frequency-driven confusions during training. In contrast, substitutions such as /i/ $\rightarrow$ /u/ in \textit{bo\textipa{S}ormin} and /e/ $\rightarrow$ /i/ in \textit{ewdili} may be influenced by variation between the consultants' pronunciation and the orthography used in the texts they were reading, for instance in the oblique stem morpheme -mu- $\sim$ -mi- or in the masculine verb form \textit{ewdi} $\sim$ \textit{iwdi}.

\paragraph{(2) Kina Rutul example}

~\\\textit{Expert:} mu\b{\textipa{\super Q}}\b{{\scriptsize\;G}}\super w\b{a\textipa{:}} \b{v}\textipa{1}s\b{e}\b{l}it\b{ }da\textipa{P}a\b{t'}ij \b{k}ar ha\b{\textipa{P}}\b{a}\b{t'}i\b{j}
\\\textit{ASR:} mu\b{q'}\super w\b{o}\textipa{\super Q}\b{w}\textipa{1}s\b{i} \b{r}i\b{td}a\textipa{P}a\b{d}i\b{q}ar ha\b{d}\b{ }i\b{ }
\\\textit{Translation:} ``They were ousting (people) from the village, they were doing things like this.''

Here, the rare phoneme {\scriptsize\;G} is confused with the much more frequent q', illustrating a typical frequency-driven substitution. Incorrect word boundary detection is also evident and occurs more frequently than in Archi, consistent with the more spontaneous and conversational nature of the Rutul recordings (see \S\ref{sec:datasets}). Some additional mismatches may be epiphenomenal: pharyngealization is prosodic rather than strictly segmental, so that its displacement from the root to the final vowel is not phonetically unmotivated and cannot be blamed on ASR. Similarly, the realization of intervocalic glottal stops (e.g., in \textit{da\textipa{P}at}) may vary due to phonetic weakening in casual speech or transcription conventions.

\section{Discussion}

Our analysis highlights the central role of frequency in phoneme recognition and connects, at a high level, to related work on frequency-driven learning in lexical processing such as \citet{murray2004serial}. Recently, \citet{heitmeier2024frequency} have shown that word frequency induces an S-shaped relationship with processing difficulty, arising from learning dynamics rather than articulatory complexity. Although their work concerns lexical representations and human reaction times, we observe a qualitatively similar sigmoid relationship between phoneme-level training frequency and ASR accuracy.

In addition, Archi and Rutul differ in their phonological inventories, primarily in the presence of palatalization in Kina Rutul and gemination in Archi (see Appendix~\ref{app:cyrlmap}). Despite these differences, the frequency–performance relationship appears consistently in both languages and in model families. This suggests that the observed sigmoid learning dynamics are not tied to a specific phonological system but may reflect more general properties of supervised phoneme learning in low-resource phonologically-complex ASR.

Furthermore, estimated sigmoid midpoints indicate that, excluding anomalous cases, phonemes begin to enter the steep learning regime at roughly $10^2$ training instances, suggesting a practical target for prioritizing phoneme coverage during data collection.

\section{Conclusion}

We present the first phoneme-level ASR analysis for Archi and Kina Rutul, based on curated speech–transcript resources consolidated into a benchmark suitable for ASR training and evaluation. We show that a simple phoneme-vocabulary adaptation with heuristic initialization yields substantial gains for wav2vec2-large-IPA under extremely low-resource conditions.

Our analyses indicate that many errors often attributed to phonological complexity are equally predictable from training frequency. While limited in scale, these findings suggest that frequency-informed perspectives may generalize across distinct phoneme inventories and offer practical guidance for data collection and evaluation in low-resource ASR. At the same time, broader validation across languages and typological settings remains an important direction for future work.

\newpage
\section*{Limitations}

This study focuses on two East Caucasian languages and relies on relatively small, manually curated datasets. Although each corpus contains on the order of one hour of transcribed speech, these resources constitute a meaningful first step for Archi and Kina Rutul---endangered languages with exceptionally rich phonological systems---for which no prior ASR benchmarks or training-ready resources existed. To our knowledge, East Caucasian languages are least likely to be present in widely used pretraining corpora (e.g., Common Voice, FLEURS) \citep{ardila-etal-2020-common, conneau2023fleurs}, as we did not identify any such languages in these datasets. The phoneme-level results for gpt-4o-transcribe, where several phonemes are not recognized at all (Figure \ref{fig:sigmoids} a \& o), are consistent with limited direct transfer, though the exact composition of pretraining data remains unknown.

Our analyses operate in an extreme low-resource regime, where phoneme-level effects become visible precisely because data are sparse. In this setting, we observe systematic relationships between phoneme-level training frequency and recognition performance. While such fine-grained effects may attenuate or change at larger data scales, they offer insight into ASR behavior at resolutions that are rarely accessible in higher-resource settings. That said, we do not claim direct transferability of these observations to other language families, data regimes, or representational units. 

In addition, our operationalization of phoneme complexity---approximated via diacritic length---serves only as a coarse proxy. Future work could incorporate more principled articulatory or typological measures, for example by considering cross-linguistic phoneme frequencies or articulatory feature inventories.

\section*{Ethics Statement}

The speech recordings and transcriptions used in this study were obtained with permission from the original authors of the respective documentation projects and were used solely for research purposes. All speakers were recorded with informed consent as part of those projects.

Large language models, including ChatGPT, were used as assistive tools during the preparation of code and manuscript text. Their use was limited to generating small, self-contained code or writing suggestions, and all such outputs were carefully reviewed, corrected, and validated by the authors. ChatGPT did not contribute original scientific ideas, analyses, or results.

\section*{Acknowledgments}
This research utilized compute resources at the Tübingen Machine Learning Cloud, DFG FKZ INST 37/1057-1 FUGG. Mahesh Akavarapu received funding from Volkswagen Foundation under the Phylomilia project within the Pioneering Projects funding line.

\bibliography{anthology,custom}

\onecolumn
\appendix
\section*{Appendix}
\section{IPA-Cyrillic map}
This section provides the mapping used to convert IPA texts into Cyrillic for employing in models like gpt-4o-transcribe or *-cyrl models. All possible phonemes occurring for each language in the datasets are also covered here. Note that a few phonemes that occur in only Russian borrowings are nevertheless included in our vocabulary and listed here. In the case of absence of one-one mapping for a phoneme due to borrowings, the native phonemes are given preference while converting from Cyrillic to IPA. 

\label{app:cyrlmap}
\begin{center}
  \includegraphics[width=\linewidth]{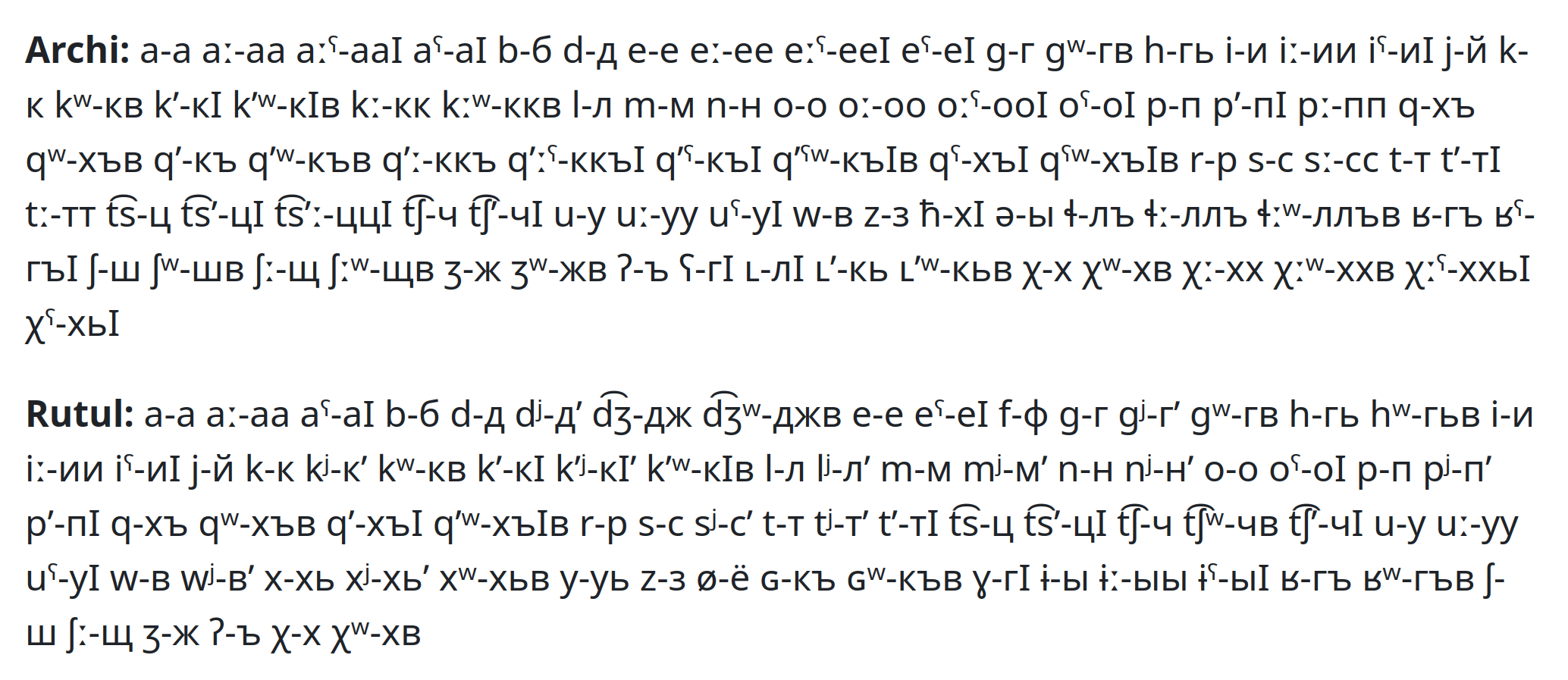}  
\end{center}

\section{Wilcoxon Signed test: p-values}
\label{app:pvals}
This section provides p-values to compare the scores of the main results (Table \ref{tab:modelres}) for each language.
\subsection{Archi}

\paragraph{Tests on CER values}
\begin{center}
\resizebox{\textwidth}{!}{
   \begin{tabular}{l|llllllll} \toprule
                                 & \textbf{\begin{tabular}[c]{@{}l@{}}w2v2l-custom\\ -avg\end{tabular}} & \textbf{\begin{tabular}[c]{@{}l@{}}w2v2l-custom\\ -avg-lm\end{tabular}} & \textbf{\begin{tabular}[c]{@{}l@{}}w2v2l-custom\\ -cpy1\end{tabular}} & \textbf{w2v2l-custom} & \textbf{\begin{tabular}[c]{@{}l@{}}wav2vec2\\ -large-ipa\end{tabular}} & \textbf{\begin{tabular}[c]{@{}l@{}}whisper\\ -large-v3\end{tabular}} & \textbf{\begin{tabular}[c]{@{}l@{}}Qwen2-Audio\\ -7B-Instruct\end{tabular}} & \textbf{\begin{tabular}[c]{@{}l@{}}Qwen2.5\\ -Omni-7B\end{tabular}} \\ \midrule
\textbf{w2v2l-custom-avg (ours)}        & -                                                                    & 0.901                                                                   & 0.810                                                                 & \textless 1e-3        & \textless 1e-3                                                         & 0.170                                                                & \textless 1e-3                                                              & \textless 1e-3                                                      \\
\textbf{w2v2l-custom-avg-lm (ours)}     & 0.901                                                                & -                                                                       & 0.847                                                                 & 0.048                 & 0.059                                                                  & 0.049                                                                & \textless 1e-3                                                              & \textless 1e-3                                                      \\
\textbf{w2v2l-custom-cpy1}       & 0.810                                                                & 0.847                                                                   & -                                                                     & 0.078                 & 0.117                                                                  & 0.151                                                                & \textless 1e-3                                                              & \textless 1e-3                                                      \\
\textbf{w2v2l-custom}            & \textless 1e-3                                                       & 0.048                                                                   & 0.078                                                                 & -                     & 0.510                                                                  & 0.001                                                                & 0.027                                                                       & \textless 1e-3                                                      \\
\textbf{wav2vec2-large-ipa}      & \textless 1e-3                                                       & 0.059                                                                   & 0.117                                                                 & 0.510                 & -                                                                      & 0.002                                                                & 0.029                                                                       & \textless 1e-3                                                      \\
\textbf{whisper-large-v3}        & 0.170                                                                & 0.049                                                                   & 0.151                                                                 & 0.001                 & 0.002                                                                  & -                                                                    & \textless 1e-3                                                              & \textless 1e-3                                                      \\
\textbf{Qwen2-Audio-7B-Instruct} & \textless 1e-3                                                       & \textless 1e-3                                                          & \textless 1e-3                                                        & 0.027                 & 0.029                                                                  & \textless 1e-3                                                       & -                                                                           & 0.002                                                               \\
\textbf{Qwen2.5-Omni-7B}         & \textless 1e-3                                                       & \textless 1e-3                                                          & \textless 1e-3                                                        & \textless 1e-3        & \textless 1e-3                                                         & \textless 1e-3                                                       & 0.002                                                                       & -   \\ \bottomrule
\end{tabular}
}
\end{center}
\paragraph{Tests on WER values}
\begin{center}
\resizebox{\textwidth}{!}{
   \begin{tabular}{l|llllllll} \toprule
                                 & \textbf{\begin{tabular}[c]{@{}l@{}}w2v2l-custom\\ -avg\end{tabular}} & \textbf{\begin{tabular}[c]{@{}l@{}}w2v2l-custom\\ -avg-lm\end{tabular}} & \textbf{\begin{tabular}[c]{@{}l@{}}w2v2l-custom\\ -cpy1\end{tabular}} & \textbf{w2v2l-custom} & \textbf{\begin{tabular}[c]{@{}l@{}}wav2vec2\\ -large-ipa\end{tabular}} & \textbf{\begin{tabular}[c]{@{}l@{}}whisper\\ -large-v3\end{tabular}} & \textbf{\begin{tabular}[c]{@{}l@{}}Qwen2-Audio\\ -7B-Instruct\end{tabular}} & \textbf{\begin{tabular}[c]{@{}l@{}}Qwen2.5\\ -Omni-7B\end{tabular}} \\ \midrule
\textbf{w2v2l-custom-avg (ours)}        & -                                                                    & 0.465                                                                   & 0.641                                                                 & \textless 1e-3        & \textless 1e-3                                                         & 0.039                                                                & 0.005                                                                       & \textless 1e-3                                                      \\
\textbf{w2v2l-custom-avg-lm (ours)}     & 0.465                                                                & -                                                                       & 0.928                                                                 & \textless 1e-3        & \textless 1e-3                                                         & 0.036                                                                & \textless 1e-3                                                              & \textless 1e-3                                                      \\
\textbf{w2v2l-custom-cpy1}       & 0.641                                                                & 0.928                                                                   & -                                                                     & \textless 1e-3        & 0.001                                                                  & 0.079                                                                & 0.002                                                                       & \textless 1e-3                                                      \\
\textbf{w2v2l-custom}            & \textless 1e-3                                                       & \textless 1e-3                                                          & \textless 1e-3                                                        & -                     & 0.891                                                                  & \textless 1e-3                                                       & 0.822                                                                       & \textless 1e-3                                                      \\
\textbf{wav2vec2-large-ipa}      & \textless 1e-3                                                       & \textless 1e-3                                                          & 0.001                                                                 & 0.891                 & -                                                                      & \textless 1e-3                                                       & 0.919                                                                       & \textless 1e-3                                                      \\
\textbf{whisper-large-v3}        & 0.039                                                                & 0.036                                                                   & 0.079                                                                 & \textless 1e-3        & \textless 1e-3                                                         & -                                                                    & \textless 1e-3                                                              & \textless 1e-3                                                      \\
\textbf{Qwen2-Audio-7B-Instruct} & 0.005                                                                & \textless 1e-3                                                          & 0.002                                                                 & 0.822                 & 0.919                                                                  & \textless 1e-3                                                       & -                                                                           & \textless 1e-3                                                      \\
\textbf{Qwen2.5-Omni-7B}         & \textless 1e-3                                                       & \textless 1e-3                                                          & \textless 1e-3                                                        & \textless 1e-3        & \textless 1e-3                                                         & \textless 1e-3                                                       & \textless 1e-3                                                              & -  \\ \bottomrule
\end{tabular}
}
\end{center}
\subsection{Kina Rutul}
\paragraph{Tests on CER values}
\begin{center}
\resizebox{\textwidth}{!}{
   \begin{tabular}{l|llllllll} \toprule
                                 & \textbf{\begin{tabular}[c]{@{}l@{}}w2v2l-custom\\ -avg\end{tabular}} & \textbf{\begin{tabular}[c]{@{}l@{}}w2v2l-custom\\ -avg-lm\end{tabular}} & \textbf{\begin{tabular}[c]{@{}l@{}}w2v2l-custom\\ -cpy1\end{tabular}} & \textbf{w2v2l-custom} & \textbf{\begin{tabular}[c]{@{}l@{}}wav2vec2\\ -large-ipa\end{tabular}} & \textbf{\begin{tabular}[c]{@{}l@{}}whisper\\ -large-v3\end{tabular}} & \textbf{\begin{tabular}[c]{@{}l@{}}Qwen2-Audio\\ -7B-Instruct\end{tabular}} & \textbf{\begin{tabular}[c]{@{}l@{}}Qwen2.5\\ -Omni-7B\end{tabular}} \\ \midrule
\textbf{w2v2l-custom-avg (ours)}        & -                                                                    & 0.638                                                                   & 0.801                                                                 & \textless 1e-3        & \textless 1e-3                                                         & 0.013                                                                & 0.003                                                                       & \textless 1e-3                                                      \\
\textbf{w2v2l-custom-avg-lm (ours)}     & 0.638                                                                & -                                                                       & 0.737                                                                 & 0.038                 & 0.067                                                                  & \textless 1e-3                                                       & \textless 1e-3                                                              & \textless 1e-3                                                      \\
\textbf{w2v2l-custom-cpy1}       & 0.801                                                                & 0.737                                                                   & -                                                                     & 0.112                 & 0.076                                                                  & 0.040                                                                & 0.010                                                                       & \textless 1e-3                                                      \\
\textbf{w2v2l-custom}            & \textless 1e-3                                                       & 0.038                                                                   & 0.112                                                                 & -                     & 0.957                                                                  & 0.458                                                                & 0.373                                                                       & 0.020                                                               \\
\textbf{wav2vec2-large-ipa}      & \textless 1e-3                                                       & 0.067                                                                   & 0.076                                                                 & 0.957                 & -                                                                      & 0.505                                                                & 0.325                                                                       & 0.012                                                               \\
\textbf{whisper-large-v3}        & 0.013                                                                & \textless 1e-3                                                          & 0.040                                                                 & 0.458                 & 0.505                                                                  & -                                                                    & 0.679                                                                       & 0.014                                                               \\
\textbf{Qwen2-Audio-7B-Instruct} & 0.003                                                                & \textless 1e-3                                                          & 0.010                                                                 & 0.373                 & 0.325                                                                  & 0.679                                                                & -                                                                           & 0.002                                                               \\
\textbf{Qwen2.5-Omni-7B}         & \textless 1e-3                                                       & \textless 1e-3                                                          & \textless 1e-3                                                        & 0.020                 & 0.012                                                                  & 0.014                                                                & 0.002                                                                       & -   \\ \bottomrule
\end{tabular}
}
\end{center}
\paragraph{Tests on WER values}
\begin{center}
\resizebox{\textwidth}{!}{
   \begin{tabular}{l|llllllll} \toprule
                                 & \textbf{\begin{tabular}[c]{@{}l@{}}w2v2l-custom\\ -avg\end{tabular}} & \textbf{\begin{tabular}[c]{@{}l@{}}w2v2l-custom\\ -avg-lm\end{tabular}} & \textbf{\begin{tabular}[c]{@{}l@{}}w2v2l-custom\\ -cpy1\end{tabular}} & \textbf{w2v2l-custom} & \textbf{\begin{tabular}[c]{@{}l@{}}wav2vec2\\ -large-ipa\end{tabular}} & \textbf{\begin{tabular}[c]{@{}l@{}}whisper\\ -large-v3\end{tabular}} & \textbf{\begin{tabular}[c]{@{}l@{}}Qwen2-Audio\\ -7B-Instruct\end{tabular}} & \textbf{\begin{tabular}[c]{@{}l@{}}Qwen2.5\\ -Omni-7B\end{tabular}} \\ \midrule
\textbf{w2v2l-custom-avg (ours)}        & -                                                                    & 0.157                                                                   & 0.706                                                                 & 0.067                 & 0.006                                                                  & 0.342                                                                & 0.107                                                                       & 0.002                                                               \\
\textbf{w2v2l-custom-avg-lm (ours)}     & 0.157                                                                & -                                                                       & 0.173                                                                 & 0.007                 & \textless 1e-3                                                         & \textless 1e-3                                                       & 0.003                                                                       & \textless 1e-3                                                      \\
\textbf{w2v2l-custom-cpy1}       & 0.706                                                                & 0.173                                                                   & -                                                                     & 0.307                 & 0.041                                                                  & 0.185                                                                & 0.135                                                                       & 0.001                                                               \\
\textbf{w2v2l-custom}            & 0.067                                                                & 0.007                                                                   & 0.307                                                                 & -                     & 0.281                                                                  & 0.747                                                                & 0.955                                                                       & 0.072                                                               \\
\textbf{wav2vec2-large-ipa}      & 0.006                                                                & \textless 1e-3                                                          & 0.041                                                                 & 0.281                 & -                                                                      & 0.487                                                                & 0.761                                                                       & 0.398                                                               \\
\textbf{whisper-large-v3}        & 0.342                                                                & \textless 1e-3                                                          & 0.185                                                                 & 0.747                 & 0.487                                                                  & -                                                                    & 0.391                                                                       & 0.013                                                               \\
\textbf{Qwen2-Audio-7B-Instruct} & 0.107                                                                & 0.003                                                                   & 0.135                                                                 & 0.955                 & 0.761                                                                  & 0.391                                                                & -                                                                           & 0.047                                                               \\
\textbf{Qwen2.5-Omni-7B}         & 0.002                                                                & \textless 1e-3                                                          & 0.001                                                                 & 0.072                 & 0.398                                                                  & 0.013                                                                & 0.047                                                                       & -   \\ \bottomrule
\end{tabular}
}
\end{center}

\section{Quality vs Quantity in Rutul}
\label{app:rutulqvsq}
In the case of Kina Rutul, the set split2 contains good quality speech (as judged by the expert, i.e., the second author), however consist only 394 sentences (utterances) while split3 contains speech of acceptable quality with 994 sentences. The evaluations on each of these splits is tabulated here here.

\begin{center}
\resizebox{0.45\textwidth}{!}{
    \begin{tabular}{l|l|lll} \toprule
\textbf{Model}     & \textbf{Split} & \textbf{WER}   & \textbf{CER}   & \textbf{PER}   \\ \midrule
w2v2l-custom-avg   & split2         & 0.891          & 0.264          & 0.252          \\
w2v2l-custom-avg   & split3         & 0.783          & \textbf{0.216} & \textbf{0.213} \\
wav2vec2-large-ipa & split2         & 0.880           & 0.271          & 0.259          \\
wav2vec2-large-ipa & split3         & \textbf{0.782} & 0.226          & 0.223    \\ \bottomrule    
\end{tabular}
}
\end{center}

\section{Phonemes with least F1 scores}
\label{app:toperr}
The top-10 most challenging phonemes for each model and language are listed here. Each triplet represents (<phoneme>, F1 score, test frequency).
\subsection{Archi}
\begin{center}
\resizebox{\textwidth}{!}{
\begin{tabular}{l|llllllllll} \toprule
\textbf{Model}               & \textbf{1}      & \textbf{2}      & \textbf{3}      & \textbf{4}      & \textbf{5}       & \textbf{6}      & \textbf{7}        & \textbf{8}        & \textbf{9}        & \textbf{10}      \\ \midrule
w2v2l-custom-avg             & (g\super w, 0.0, 6)  & (i\textipa{:}, 0.0, 6)  & (\textipa{X}\textipa{\super Q}, 0.0, 5)  & (a\textipa{\super Q}, 0.0, 4)  & (o\textipa{:}\textipa{\super Q}, 0.0, 4)  & (i\textipa{\super Q}, 0.0, 3)  & (\textipa{\textbeltl}\textipa{:}\super w, 0.0, 3)   & (o\textipa{\super Q}, 0.0, 2)    & (q'\super w, 0.0, 2)   & (q'\textipa{:}\textipa{\super Q}, 0.0, 2) \\
w2v2l-custom-avg-lm          & (g\super w, 0.0, 6)  & (i\textipa{:}, 0.0, 6)  & (\textipa{X}\textipa{\super Q}, 0.0, 5)  & (a\textipa{\super Q}, 0.0, 4)  & (o\textipa{:}\textipa{\super Q}, 0.0, 4)  & (i\textipa{\super Q}, 0.0, 3)  & (\textipa{\textbeltl}\textipa{:}\super w, 0.0, 3)   & (o\textipa{\super Q}, 0.0, 2)    & (q'\super w, 0.0, 2)   & (q'\textipa{:}\textipa{\super Q}, 0.0, 2) \\
w2v2l-custom-cpy1            & (g\super w, 0.0, 6)  & (i\textipa{:}, 0.0, 6)  & (\textipa{X}\textipa{\super Q}, 0.0, 5)  & (a\textipa{\super Q}, 0.0, 4)  & (i\textipa{\super Q}, 0.0, 3)   & (q'\super w, 0.0, 2) & (q'\textipa{\super Q}, 0.0, 2)   & (k\super w, 0.0, 1)    & (\textipa{\t{ts}}'\textipa{:}, 0.0, 1) & (x\textipa{:}\textipa{\super Q}, 0.0, 1)  \\
w2v2l-custom                 & (e\textipa{\super Q}, 0.0, 8)  & (g\super w, 0.0, 6)  & (i\textipa{:}, 0.0, 6)  & (\textipa{\t{ts}}, 0.0, 6) & (\textipa{@}, 0.0, 6)    & (q'\textipa{:}, 0.0, 5) & (\textipa{X}\textipa{\super Q}, 0.0, 5)    & (a\textipa{\super Q}, 0.0, 4)    & (o\textipa{:}\textipa{\super Q}, 0.0, 4)   & (i\textipa{\super Q}, 0.0, 3)   \\
w2v2l-custom-avg-zs          & (t\textipa{:}, 0.0, 89) & (\textipa{X}, 0.0, 78)  & (t', 0.0, 57) & (\textipa{\textbeltl}\textipa{:}, 0.0, 37) & (q\textipa{\super Q}, 0.0, 23)  & (q, 0.0, 21)  & (k', 0.0, 17)   & (\;L', 0.0, 15)   & (\textipa{\t{t\textipa{S}}}', 0.0, 14) & (\textipa{\textbeltl}, 0.0, 12)   \\
gpt-4o-transcribe            & (h, 0.0, 42)  & (\textipa{\textbeltl}\textipa{:}, 0.0, 37) & (q\textipa{\super Q}, 0.0, 23) & (q, 0.0, 21)  & (k', 0.0, 17)  & (\;L', 0.0, 15) & (\textipa{\t{t\textipa{S}}}', 0.0, 14) & (\textipa{\textbeltl}, 0.0, 12)    & (\textipa{\t{ts}}', 0.0, 10) & (e\textipa{\super Q}, 0.0, 8)   \\
Qwen2-Audio-7B-Instruct      & (e\textipa{\super Q}, 0.0, 8)  & (g\super w, 0.0, 6)  & (i\textipa{:}, 0.0, 6)  & (\textipa{@}, 0.0, 6)   & (\textipa{X}\textipa{\super Q}, 0.0, 5)   & (a\textipa{\super Q}, 0.0, 4)  & (o\textipa{:}\textipa{\super Q}, 0.0, 4)   & (i\textipa{\super Q}, 0.0, 3)    & (\textipa{\textbeltl}\textipa{:}\super w, 0.0, 3)   & (k'\super w, 0.0, 2)  \\
Qwen2-Audio-7B-Instruct-cyrl & (\textipa{@}, 0.0, 6)   & (\textipa{X}\textipa{\super Q}, 0.0, 5)  & (a\textipa{\super Q}, 0.0, 4)  & (i\textipa{\super Q}, 0.0, 3)  & (\textipa{\textipa{S}}\textipa{:}, 0.0, 3)   & (q'\super w, 0.0, 2) & (q'\textipa{:}\textipa{\super Q}, 0.0, 2)  & (\textipa{Z}\super w, 0.0, 2)    & (\textipa{\textipa{S}}\textipa{:}\super w, 0.0, 1)   & (\textipa{X}\textipa{:}\super w, 0.0, 1)  \\
Qwen2.5-Omni-7B              & (e\textipa{\super Q}, 0.0, 8)  & (\;L, 0.0, 7)   & (\textipa{X}\textipa{:}\textipa{\super Q}, 0.0, 7) & (g\super w, 0.0, 6)  & (\textipa{@}, 0.0, 6)    & (\textipa{X}\textipa{\super Q}, 0.0, 5)  & (a\textipa{\super Q}, 0.0, 4)    & (o\textipa{:}\textipa{\super Q}, 0.0, 4)   & (i\textipa{\super Q}, 0.0, 3)    & (q'\textipa{\super {Qw}}, 0.0, 3) \\
Qwen2.5-Omni-7B-cyrl         & (k', 0.0, 17) & (\;L', 0.0, 15) & (e\textipa{\super Q}, 0.0, 8)  & (\textipa{X}\textipa{:}\textipa{\super Q}, 0.0, 7) & (\textipa{@}, 0.0, 6)    & (q'\textipa{:}, 0.0, 5) & (\textipa{\textcrh}, 0.0, 5)     & (\textipa{X}\textipa{\super Q}, 0.0, 5)    & (a\textipa{\super Q}, 0.0, 4)    & (o\textipa{:}\textipa{\super Q}, 0.0, 4)  \\
wav2vec2-large-ipa           & (i\textipa{:}, 0.0, 6)  & (a\textipa{\super Q}, 0.0, 4)  & (i\textipa{\super Q}, 0.0, 3)  & (q'\super w, 0.0, 2) & (q'\textipa{:}\textipa{\super Q}, 0.0, 2) & (\textipa{Z}\super w, 0.0, 2)  & (\textipa{\t{ts}}'\textipa{:}, 0.0, 1) & (x\textipa{:}\textipa{\super Q}, 0.0, 1)   & (\textipa{@}, 0.15, 6)    & (q'\textipa{\super Q}, 0.33, 2) \\
wav2vec2-large-ipa-zs        & (\textipa{X}, 0.0, 78)  & (t', 0.0, 57) & (\textipa{\textbeltl}\textipa{:}, 0.0, 37) & (q\textipa{\super Q}, 0.0, 23) & (q, 0.0, 21)   & (k', 0.0, 17) & (\;L', 0.0, 15)   & (\textipa{\t{t\textipa{S}}}', 0.0, 14) & (\textipa{\textbeltl}, 0.0, 12)    & (s\textipa{:}, 0.0, 10)  \\
whisper-large-v3             & (\textipa{@}, 0.0, 6)   & (a\textipa{\super Q}, 0.0, 4)  & (i\textipa{\super Q}, 0.0, 3)  & (k'\super w, 0.0, 2) & (q'\textipa{:}\textipa{\super Q}, 0.0, 2) & (q'\textipa{\super Q}, 0.0, 2) & (k\super w, 0.0, 1)    & (\textipa{\t{ts}}'\textipa{:}, 0.0, 1) & (i\textipa{:}, 0.5, 6)    & (\textipa{K}, 0.5, 4)    \\
whisper-large-v3-cyrl        & (\textipa{@}, 0.0, 6)   & (\textipa{\textcrh}, 0.0, 5)   & (\textipa{X}\textipa{\super Q}, 0.0, 5)  & (i\textipa{\super Q}, 0.0, 3)  & (q'\textipa{:}\textipa{\super Q}, 0.0, 2) & (q'\textipa{\super Q}, 0.0, 2) & (\textipa{Q}, 0.06, 4)    & (i\textipa{:}, 0.29, 6)   & (a\textipa{\super Q}, 0.33, 4)   & (k', 0.54, 17) \\ \bottomrule
\end{tabular} 
}
\end{center}

\subsection{Kina Rutul}
\begin{center}
\resizebox{\textwidth}{!}{
\begin{tabular}{l|llllllllll} \toprule
\textbf{Model}               & \textbf{1}      & \textbf{2}      & \textbf{3}      & \textbf{4}      & \textbf{5}      & \textbf{6}      & \textbf{7}       & \textbf{8}       & \textbf{9}       & \textbf{10}      \\ \midrule
w2v2l-custom-avg             & (k'\super j, 0.0, 6) & (l\super j, 0.0, 5)  & (q'\super w, 0.0, 2) & (u\textipa{:}, 0.0, 2)  & (\textipa{\textipa{S}}j\textipa{:}, 0.0, 2) & (d\super j, 0.0, 1)  & (p', 0.0, 1)   & (\;G\super w, 0.0, 1)   & (\textipa{G}, 0.0, 1)    & (\textipa{1}\textipa{\super Q}, 0.0, 1)   \\
w2v2l-custom-avg-lm          & (k'\super j, 0.0, 6) & (q'\super w, 0.0, 2) & (u\textipa{:}, 0.0, 2)  & (\textipa{\textipa{S}}\super j\textipa{:}, 0.0, 2) & (d\super j, 0.0, 1)  & (p', 0.0, 1)  & (\;G\super w, 0.0, 1)   & (\textipa{G}, 0.0, 1)    & (\textipa{1}\textipa{\super Q}, 0.0, 1)   & (a\textipa{:}, 0.15, 25) \\
w2v2l-custom-cpy1            & (k'\super j, 0.0, 6) & (u\textipa{:}, 0.0, 2)  & (\textipa{\textipa{S}}\super j\textipa{:}, 0.0, 2) & (d\super j, 0.0, 1)  & (k', 0.0, 1)  & (p', 0.0, 1)  & (\;G\super w, 0.0, 1)   & (\textipa{1}\textipa{\super Q}, 0.0, 1)   & (o, 0.1, 35)   & (v, 0.22, 10)  \\
w2v2l-custom                 & (v, 0.0, 10)  & (k'\super j, 0.0, 6) & (\textipa{\t{ts}}, 0.0, 6) & (l\super j, 0.0, 5)  & (g\super w, 0.0, 3)  & (k\super w, 0.0, 3)  & (q'\super w, 0.0, 2)  & (u\textipa{:}, 0.0, 2)   & (\textipa{\textipa{S}}\super j\textipa{:}, 0.0, 2)  & (d\super j, 0.0, 1)   \\
w2v2l-custom-avg-zs          & (\textipa{X}, 0.0, 72)  & (x\super w, 0.0, 58) & (\textipa{K}, 0.0, 45)  & (\;G, 0.0, 32)  & (q', 0.0, 27) & (a\textipa{:}, 0.0, 25) & (a\textipa{\super Q}, 0.0, 21)  & (q, 0.0, 21)   & (u\textipa{\super Q}, 0.0, 13)  & (y, 0.0, 12)   \\
gpt-4o-transcribe            & (h, 0.0, 139) & (\textipa{Q}, 0.0, 89)  & (x\super w, 0.0, 58) & (\textipa{K}, 0.0, 45)  & (q', 0.0, 27) & (a\textipa{\super Q}, 0.0, 21) & (q, 0.0, 21)   & (u\textipa{\super Q}, 0.0, 13)  & (y, 0.0, 12)   & (t', 0.0, 10)  \\
Qwen2-Audio-7B-Instruct      & (k'\super j, 0.0, 6) & (g\super w, 0.0, 3)  & (q'\super w, 0.0, 2) & (u\textipa{:}, 0.0, 2)  & (\textipa{\textipa{S}}\super j\textipa{:}, 0.0, 2) & (k', 0.0, 1)  & (p', 0.0, 1)   & (\textipa{G}, 0.0, 1)    & (\textipa{1}\textipa{\super Q}, 0.0, 1)   & (u\textipa{\super Q}, 0.11, 13) \\
Qwen2-Audio-7B-Instruct-cyrl & (\textipa{Q}, 0.0, 89)  & (q', 0.0, 27) & (a\textipa{\super Q}, 0.0, 21) & (u\textipa{\super Q}, 0.0, 13) & (t', 0.0, 10) & (v, 0.0, 10)  & (\textipa{\t{ts}}', 0.0, 8) & (\textipa{\t{t\textipa{S}}}', 0.0, 8) & (k'\super j, 0.0, 6)  & (l\super j, 0.0, 5)   \\
Qwen2.5-Omni-7B              & (u\textipa{\super Q}, 0.0, 13) & (k'\super j, 0.0, 6) & (q'\super w, 0.0, 2) & (u\textipa{:}, 0.0, 2)  & (\textipa{\textipa{S}}\super j\textipa{:}, 0.0, 2) & (d\super j, 0.0, 1)  & (k', 0.0, 1)   & (p', 0.0, 1)   & (\;G\super w, 0.0, 1)   & (\textipa{G}, 0.0, 1)    \\
Qwen2.5-Omni-7B-cyrl         & (\textipa{Q}, 0.0, 89)  & (q', 0.0, 27) & (a\textipa{\super Q}, 0.0, 21) & (u\textipa{\super Q}, 0.0, 13) & (y, 0.0, 12)  & (t', 0.0, 10) & (v, 0.0, 10)   & (\textipa{\t{ts}}', 0.0, 8) & (\textipa{\t{t\textipa{S}}}', 0.0, 8) & (k'\super j, 0.0, 6)  \\
wav2vec2-large-ipa           & (k'\super j, 0.0, 6) & (q'\super w, 0.0, 2) & (u\textipa{:}, 0.0, 2)  & (\textipa{\textipa{S}}\super j\textipa{:}, 0.0, 2) & (d\super j, 0.0, 1)  & (p', 0.0, 1)  & (\;G\super w, 0.0, 1)   & (\textipa{1}\textipa{\super Q}, 0.0, 1)   & (a\textipa{:}, 0.08, 25) & (o, 0.12, 35)  \\
wav2vec2-large-ipa-zs        & (\textipa{X}, 0.0, 72)  & (x\super w, 0.0, 58) & (\textipa{K}, 0.0, 45)  & (\;G, 0.0, 32)  & (q', 0.0, 27) & (a\textipa{\super Q}, 0.0, 21) & (q, 0.0, 21)   & (u\textipa{\super Q}, 0.0, 13)  & (y, 0.0, 12)   & (t', 0.0, 10)  \\
whisper-large-v3             & (t', 0.0, 10) & (k'\super j, 0.0, 6) & (k\super w, 0.0, 3)  & (q'\super w, 0.0, 2) & (u\textipa{:}, 0.0, 2)  & (\textipa{\textipa{S}}\super j\textipa{:}, 0.0, 2) & (d\super j, 0.0, 1)   & (k', 0.0, 1)   & (p', 0.0, 1)   & (\;G\super w, 0.0, 1)   \\
whisper-large-v3-cyrl        & (j, 0.0, 89)  & (q', 0.0, 27) & (a\textipa{\super Q}, 0.0, 21) & (u\textipa{\super Q}, 0.0, 13) & (t', 0.0, 10) & (v, 0.0, 10)  & (\textipa{\t{ts}}', 0.0, 8) & (\textipa{\t{t\textipa{S}}}', 0.0, 8) & (k'\super j, 0.0, 6)  & (l\super j, 0.0, 5) \\ \bottomrule 
\end{tabular}
}
\end{center}

\end{document}